\documentclass[times, review, 10pt]{elsarticle}
\usepackage{lineno,hyperref}
\usepackage[subrefformat=parens,labelformat=parens]{subfig}

\usepackage{soul}

\usepackage{tabularx}

\modulolinenumbers[5]

\usepackage{amssymb}
\usepackage{amsmath}
\usepackage{dsfont}

\usepackage{rotating}
\usepackage{booktabs}
\usepackage{multirow}
\usepackage{comment}
\usepackage{pifont}

\usepackage{setspace}
\journal{Pattern Recognition}

\begin{document}

\begin{frontmatter}



\title{ASV3D: Adapting Diffusion-Based Single-View 3D Reconstruction with Extra Imagery}


\author[label1]{Y Huynh}
\author[label2]{Duc Thanh Nguyen\corref{cor1}}
\author[label3]{Thao Minh Le}
\author[label1]{Mohamed Abdelrazek}

\affiliation[label1]{organization={Applied Artificial Intelligence Initiative},
            country={Australia}}
\affiliation[label2]{organization={School of Information Technology, Deakin University},
            country={Australia}}
\affiliation[label3]{organization={Pennsylvania State University}, country={The United States of America}}

\ead{duc.nguyen@deakin.edu.au}
\cortext[cor1]{Corresponding author.}

\begin{abstract}

Reconstruction of 3D objects from a single image is a fundamental research topic in computer vision. The key challenge is the lack of information from critical viewpoints to complete 3D structures. Using an additional view may help to resolve the issue. However, there is no mechanism that can integrate the extra view into the diffusion-based single-view 3D reconstruction principle. We address this challenge by proposing \textbf{ASV3D}, a framework for adapting diffusion-based single-view 3D object reconstruction to test-time data with support from one additional image. We introduce two adaptation strategies: (i) a \textit{zero-shot adaptation} scheme that leverages the auxiliary image to improve the reconstruction quality of an object without retraining, and (ii) an \textit{optimised adaptation} scheme that further enhances visual fidelity and cross-view consistency via contrastive learning. We apply our ASV3D to improve two state-of-the-art diffusion-based single-view 3D reconstruction pipelines on both benchmark and real-world datasets. Results demonstrate that our approach consistently improves reconstruction accuracy and robustness under unconstrained multi-view inputs, outperforming the baselines in both quantitative metrics and human preference. We publish our code and the real-world object dataset on our project page at \url{https://github.com/YNhuHuynh/ASV3D/tree/main}.

\end{abstract}




\end{frontmatter}



\section{Introduction}
\label{sec:Introduction}

Single-view 3D reconstruction has long been a fundamental problem in computer vision. Its goal is to recover a complete 3D representation of an object from a single image. Early approaches relied on geometric cues, such as depth, to reconstruct object surfaces~\cite{DBLP:conf/nips/SaxenaCN05, DBLP:conf/cvpr/OswaldTC12}. However, these methods are inherently limited to reconstructing only visible portions of an object. To infer invisible or occluded regions, learning-based approaches exploit structural priors learned from large-scale data~\cite{DBLP:journals/pami/TulsianiKCM17, DBLP:conf/eccv/ChoyXGCS16, DBLP:conf/eccv/GirdharFRG16, DBLP:conf/nips/0001ZXFT16, DBLP:conf/iccv/TatarchenkoDB17, DBLP:conf/cvpr/GroueixFKRA18, DBLP:conf/cvpr/Richter018}. 

Recent advances in generative artificial intelligence have significantly advanced single-view 3D reconstruction. In particular, diffusion models~\cite{DBLP:conf/cvpr/RombachBLEO22, ruiz_dreambooth_2023, DBLP:conf/cvpr/ShumHNY25} have demonstrated remarkable capability in synthesising high-quality, photo-realistic images from simple user inputs, such as text prompts or reference images. Leveraging this capability, a new class of methods, referred to as \textit{diffusion-based conditional generative 3D reconstruction}, first synthesises consistent novel views conditioned on the input image using a diffusion model and then reconstructs a high-fidelity 3D model from the generated views~\cite{long_wonder3d_2024, DBLP:conf/nips/LiLLZLLQZXLTWLG24}.

\begin{figure}
    \centering
    \small
    \includegraphics[width=1.0\linewidth]{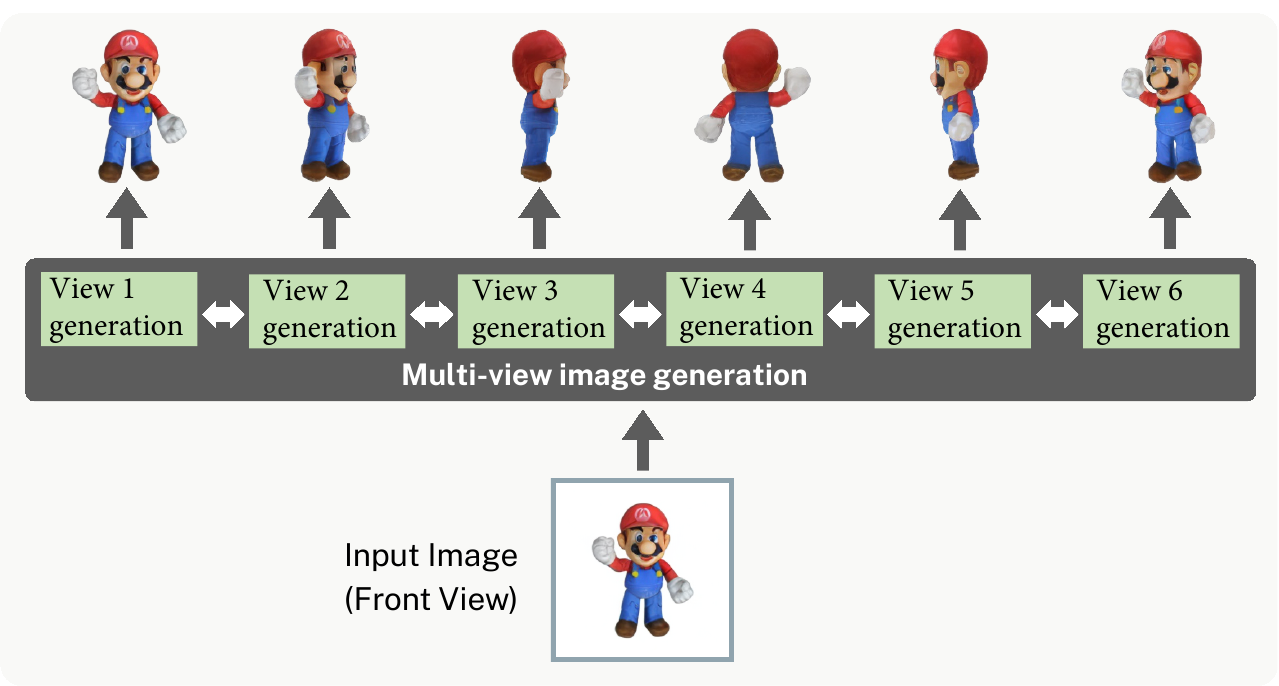}\\
    \text{(a)}\\
    \includegraphics[width=1.0\linewidth]{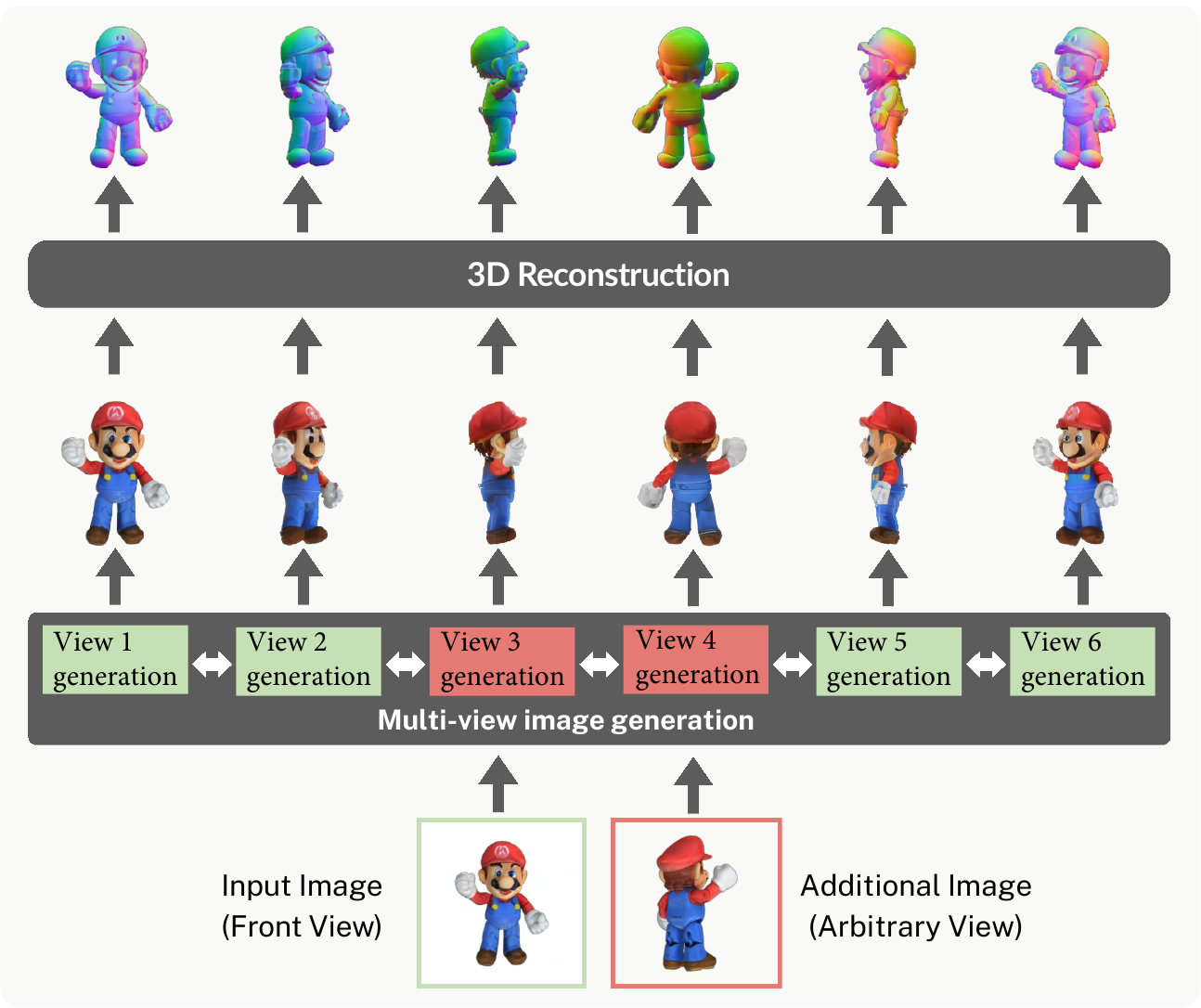}\\
    \text{(b)}
    \caption{\textbf{ASV3D vs. Wonder3D}. (a) Wonder3D~\cite{long_wonder3d_2024} uses the same input image to condition the generation of all target views. (b) Our ASV3D drives the input and additional images into proper generations, highlighted in corresponding colours (green:  input image, red: additional one).}
    \label{fig:teaser}
\end{figure}

Despite their impressive performance, existing diffusion-based conditional generative methods often struggle to reconstruct unfamiliar objects at test time. The generated novel views frequently exhibit geometric misalignment and cross-view inconsistencies, which subsequently degrade the quality of the reconstructed 3D model. The primary reason is the extremely under-constrained nature of the single-image setting. For example, synthesising the rear view of an unfamiliar object from only its frontal image is highly ambiguous because little or no visual evidence is available for the unseen regions.

To alleviate this limitation, we propose exploiting an additional image of the same object to supplement the missing information. The key challenge, however, is determining how to effectively utilise the additional image in the diffusion-based single-view setting. Existing multi-conditioning methods~\cite{DBLP:conf/cvpr/YangHWGZZ0L24, DBLP:conf/cvpr/CaoYL0X025} typically fuse all input images into the condition of a diffusion model to synthesise every target view. Such a unified conditioning strategy often produces suboptimal results, particularly for distant or heavily occluded surfaces. Different target views require different visual evidence: some input images provide essential information for a particular view, whereas others may be redundant or even introduce ambiguity. Motivated by this observation, we propose a consistency-based gate that automatically identifies the most suitable conditioning image for each target view. As illustrated in Figure~\ref{fig:teaser}, selectively conditioning each generated view with the most informative input substantially improves multi-view generation and ultimately leads to more accurate 3D reconstruction.

Our problem setting differs fundamentally from the sparse-view 3D reconstruction~\cite{DBLP:conf/cvpr/KongLLT0D24, zhao2024sparseags, Xu_2025_ICCV}. Firstly, our framework accepts a primary image and an additional image captured by different cameras. Secondly, it does not require pose estimation, as opposed to several unposed sparse-view 3D reconstruction methods, e.g.,~\cite{DBLP:conf/cvpr/ZhangWX000SW25, DBLP:conf/iclr/YeLXLP0P25}. Finally, our framework progressively adapts the conditioning information for each target view rather than jointly processing all input images, making the reconstruction adaptive to the availability of observed data. In summary, we make the following contributions in our paper.

\begin{itemize}
\item We propose \textbf{ASV3D}, a framework that adapts diffusion-based single-view generative 3D reconstruction models using a primary image together with an additional image. The two images may be captured under different conditions, and neither camera poses nor calibration information are required. To the best of our knowledge, our work is the first to adapt diffusion-based single-view 3D reconstruction using additional imagery.

\item We introduce two adaptation strategies: \emph{zero-shot adaptation} and \emph{optimised adaptation}. The zero-shot variant uses a consistency-based gating mechanism, which automatically associates each target view with its most informative conditioning image. The optimised variant further improves cross-view consistency via contrastive learning.

\item We conducted extensive evaluations and a user study, demonstrating that ASV3D consistently improves state-of-the-art diffusion-based single-view 3D reconstruction baselines.

\end{itemize}

\section{Related Work}

\paragraph{Diffusion-based 3D reconstruction.}
Diffusion models~\cite{ho_denoising_2020,DBLP:conf/cvpr/RombachBLEO22,saharia_photorealistic_2022}, underpin recent progress in high-fidelity 3D generation~\cite{raj_dreambooth3d_2023, RealFusion, nichol2022pointe, jun2023shape, lin_magic3d_2023_1, chen2023fantasia3d, liu_one-2-3-45_2023}. Beyond text prompts, recent work shows that diffusion models can be effectively guided by visual prompts. IP-Adapter~\cite{DBLP:journals/corr/abs-2308-06721} injects reference-image features into a diffusion model via lightweight adapters attached to attention layers, enabling image-prompting without retraining the full backbone. ControlNet~\cite{DBLP:conf/iccv/ZhangRA23} conditions a diffusion model through dedicated control branches, providing a general mechanism to incorporate additional visual guidance signals. These advances enable diffusion-based single view-conditioned generators such as Zero-1-to-3~\cite{DBLP:journals/corr/abs-2303-11328}, SyncDreamer~\cite{liu_syncdreamer_2024},  Wonder3D~\cite{long_wonder3d_2024}, Era3D~\cite{DBLP:conf/nips/LiLLZLLQZXLTWLG24}, which adopt diffusion models to condition the generation of multiple views of an object for downstream reconstruction. Several works focus on the scalability and efficiency of reconstruction models, e.g., LRM~\cite{hong2023lrm}, CRM~\cite{wang2024crm}, InstantMesh~\cite{xu2024instantmesh}. Recent generative multi-view synthesis and 3D reconstruction
methods use more complex conditions, including various types of information, e.g., camera pose~\cite{DBLP:conf/cvpr/YangHWGZZ0L24} and depth~\cite{DBLP:conf/cvpr/CaoYL0X025}. However, simply adding more inputs does not necessarily improve the consistency of multi-view generation. This motivates conditioning strategies that explicitly account for heterogeneous informativeness across viewpoints.

\paragraph{Adaptation for 3D reconstruction.} Recent studies have explored test-time adaptation (TTA) to improve the generalisation of 3D reconstruction models without retraining. REFINE~\cite{DBLP:conf/cvpr/LeungHV22} performs instance-specific optimisation on the reconstructed shape using differentiable rendering and geometric regularisation while treating the underlying reconstruction model as a black box. More recent methods, such as MeTTA~\cite{DBLP:conf/bmvc/Yu-JiHYSHO24} and subsequent diffusion-based TTA approaches~\cite{DBLP:conf/nips/YuanSWYW25, DBLP:journals/corr/abs-2509-26645}, further exploit generative priors and optimise latent representations, neural fields, or camera parameters during inference to improve reconstruction quality and robustness to unseen objects and domains. Despite advances, existing TTA methods for diffusion-based single-view 3D reconstruction can only enforce view consistency (via virtual camera learning) with visible parts given in the single input image during adaptation, while the geometry of occluded or invisible regions remains unconstrained. Although powerful generative priors enable these methods to hallucinate plausible missing surfaces, the generated geometry is often inaccurate or inconsistent with the actual object, particularly for unfamiliar categories or complex shapes. This limitation stems from the inherent ambiguity of diffusion-based single-view reconstruction, where multiple 3D shapes can explain the same 2D observation.

\section{Proposed Method}

\begin{figure}[ht]
  \centering
  \includegraphics[width=1.0\linewidth]{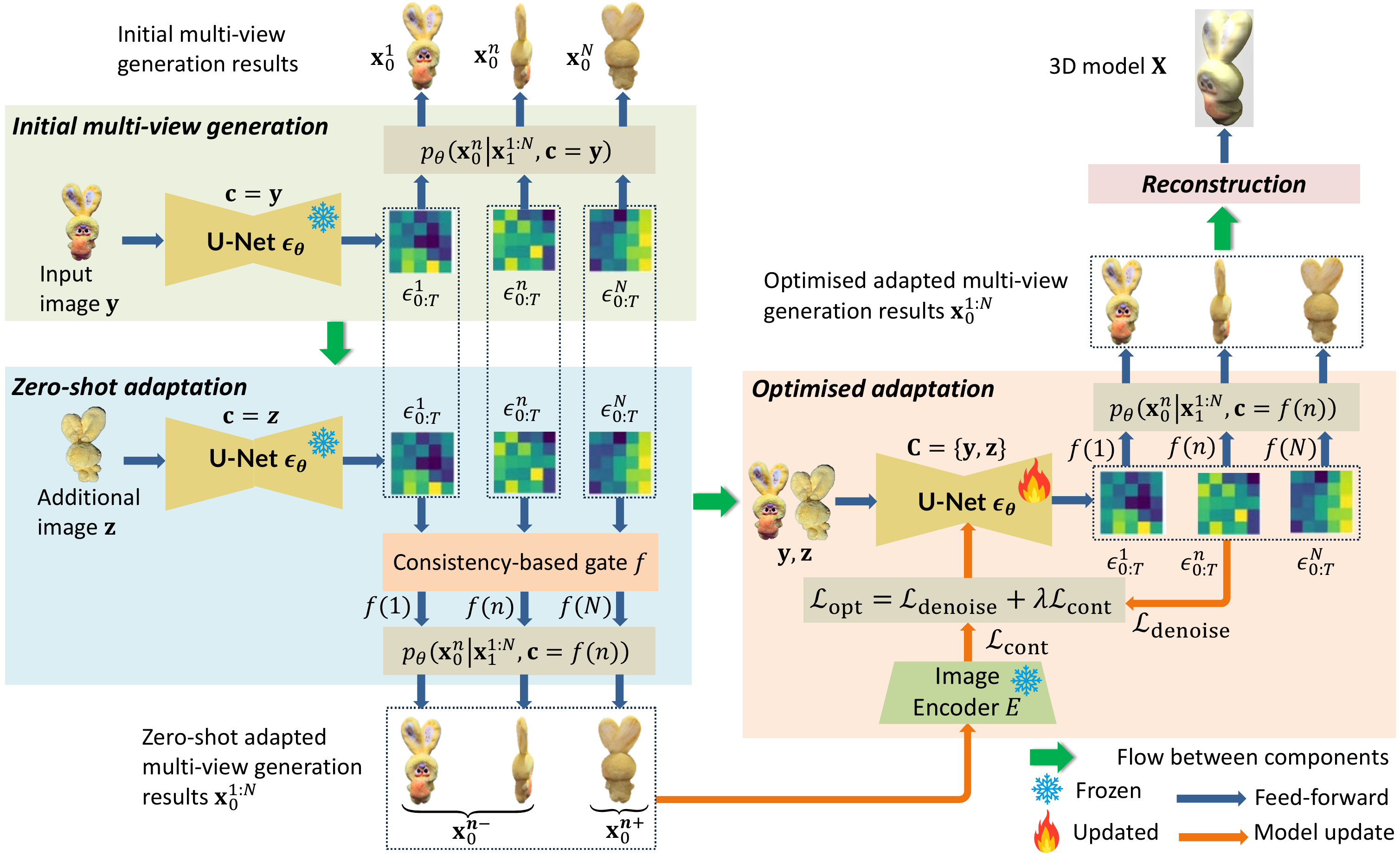}
  \caption{\textbf{Overview of ASV3D}. Given an input image $\mathbf{y}$, we apply a pre-trained generative model (e.g., Wonder3D) to generate an initial set of target views. We then adapt the model to an additional image $\mathbf{z}$ once supplied. Our method supports two adaptation strategies: zero-shot and optimised adaptation. The zero-shot setting determines the optimal condition for each target view using a consistency-based gate. The optimised adaptation further enforces multi-view consistency with contrastive learning.}
  \label{fig:overview}
\end{figure}

\subsection{Overview}

We illustrate our method in Figure~\ref{fig:overview}. Our ASV3D is built upon the principle of diffusion-based single-view 3D reconstruction, which takes as input a single image $\mathbf{y}$ of an object and reconstructs the 3D model $\mathbf{X}$ of the object. Following the standard diffusion-based single-view 3D reconstruction pipeline~\cite{long_wonder3d_2024}, we first generate $N$ views $\mathbf{x}^{1:N}$ from $\mathbf{y}$. This process is referred to as ``multi-view generation'' and can be performed by applying a diffusion-based conditional multi-view synthesis model~\cite{DBLP:conf/cvpr/RombachBLEO22} with condition $\mathbf{c}=\mathbf{y}$,
\begin{align}
    &p_{\theta}\big(\mathbf{x}_{0:T}^{1:N}|\mathbf{c}\big)=p\big(\mathbf{x}_{T}^{1:N}|\mathbf{c}\big)\prod_{t=1}^T \prod_{n=1}^N p_{\theta}\big(\mathbf{x}_{t-1}^{n}|\mathbf{x}_t^{1:N},\mathbf{c}\big) \label{eq:general_condition_1}\\
    &p_{\theta}\big(\mathbf{x}_{t-1}^{n}|\mathbf{x}_t^{1:N},\mathbf{c}\big) = \mathcal{N}\big( \mathbf{x}_{t-1}^{n};\mu_{\theta}^{n}(\mathbf{x}_{t}^{1:N},t,\mathbf{c}),\sigma_t^2 \mathbf{I} \big) \label{eq:general_condition_2}\\
    &\mu_{\theta}^{n}\big(\mathbf{x}_{t}^{1:N},t,\mathbf{c}\big)=\frac{1}{\sqrt{\alpha_t}} \bigg( \mathbf{x}_t^{n} - \frac{\beta_t}{\sqrt{1-\bar{\alpha}_t}} \epsilon_{\theta}^{n}\big(\mathbf{x}_t^{1:N},t,\mathbf{c} \big) \bigg)
    \label{eq:general_condition_3}
\end{align}
where $T$ is the number of diffusion steps, $\beta_t$ is a schedule, $\alpha_t=1-\beta_t$, $\bar{\alpha}_t=\prod_{t'=1}^t \alpha_{t'}$, $\sigma_t$ is a time-dependent variable (see details in~\cite{ho_denoising_2020}), and $\epsilon_{\theta}$ with learnable parameters $\theta$ is a U-Net estimating the noise $\epsilon^{1:N}$ added to all $N$ views. The generated views $\mathbf{x}_{0}^{1:N}$ are used to reconstruct the object $\mathbf{X}$ using the method in~\cite{DBLP:conf/nips/WangLLTKW21}.

We observed that the generation process in Eq.~(\ref{eq:general_condition_1})-(\ref{eq:general_condition_3}) often produces hallucination errors, especially for views distant from $\mathbf{y}$ due to limited evidence for reliable synthesis. For instance, the hat and hair of the object in Figure~\ref{fig:teaser}(a) are well separated in the 1\textit{st}, 2\textit{nd}, and 6\textit{th} (from left to right) generated images, which are close to the input image, but they are merged in the 3\textit{rd}, 4\textit{th}, and 5\textit{th} images. An additional image, even if captured by a different camera, can provide auxiliary information to disambiguate distant views, thereby improving cross-view consistency and fidelity of the later reconstructed 3D model. We illustrate this observation in Figure~\ref{fig:teaser}(b) where the additional image shows its impact on the multi-view image generation process. 

In this paper, we propose to adapt $\epsilon_\theta$ using an additional image $\mathbf{z}$. Unlike existing multi-view and sparse-view reconstruction methods, we allow both $\mathbf{y}$ and $\mathbf{z}$ to be captured using different cameras. The additional image $\mathbf{z}$ serves as an auxiliary information that helps to disambiguate the generation of difficult views. The key of our setting is to determine the condition $\mathbf{c}$ used in $\epsilon_{\theta}^{n}\big(\mathbf{x}_{t}^{1:N},t,\mathbf{c}\big)$, which takes a single conditioning image. The straightforward approach is to choose the condition in the generation of $\mathbf{x}^n$ as the input image (either $\mathbf{y}$ or $\mathbf{z}$) closest to $\mathbf{x}^n$. Nevertheless, this requires the estimation of the camera poses of the input images, which is also challenging because both $\mathbf{y}$ and $\mathbf{z}$ can be captured with different cameras. 

\paragraph{Consistency-based gating.} To address the above issue, we propose a camera-pose-free gating mechanism for optimal selection of conditions in multi-view generation. Our gating mechanism is designed on the basis of consistency models, measuring the consistency of a diffusion model from its denoising convergence~\cite{DBLP:conf/icml/SongD0S23, DBLP:conf/nips/DarasDDD23}. Specifically, our gate is formulated as a function $f:\{1,...,N\}\xrightarrow[]{}\mathbf{C}=\{\mathbf{y},\mathbf{z}\}$ that determines a condition $\mathbf{c}\in\mathbf{C}$, which minimises the deviation of $p_\theta(\mathbf{x}_{t-1}^n|\mathbf{x}_t^{1:N},\mathbf{c})$ through time steps $t$, for each target view index $n\in\{1,...,N\}$. Mathematically, we define
\begin{align}
    f(n)=\underset{\mathbf{c\in\mathbf{C}}}{\operatorname{argmin}} \mathbb{E}_{t\in\mathcal{T}}\bigg\{\big\|\mu_{\theta}^{n}\big(\mathbf{x}_{t}^{1:N},t,\mathbf{c}\big)-\hat{\mu}\big\|_2^2\bigg\}
    \label{eq:condition_selection}
\end{align}
where $\hat{\mu}=\mathbb{E}_{t\in\mathcal{T}}\big\{\mu_{\theta}^{n}\big(\mathbf{x}_{t}^{1:N},t,\mathbf{c}\big)\big\}$ and $\mathcal{T}=\{\frac{T}{2},...,1\}$ (late half). We empirically found that such late diffusion steps $\mathcal{T}$ perform best for condition selection.

As shown in Eq.~(\ref{eq:condition_selection}), the gate $f$ aims to choose the condition $\mathbf{c}$ that the generative model is most consistent with the generation of $\mathbf{x}_0^{n}$. Given $f$, we derive the diffusion-based conditional multi-view generation process as follows,
\begin{align}
    &p_{\theta}\big(\mathbf{x}_{0:T}^{1:N}|\mathbf{C}\big)=p\big(\mathbf{x}_{T}^{1:N}|\mathbf{C}\big)\prod_{t=1}^T \prod_{n=1}^N p_{\theta}\big(\mathbf{x}_{t-1}^{n}|\mathbf{x}_t^{1:N},f(n)\big) \label{eq:opo_1}\\
    &p_{\theta}\big(\mathbf{x}_{t-1}^{n}|\mathbf{x}_t^{1:N},f(n)\big) = \mathcal{N}\big( \mathbf{x}_{t-1}^{n};\mu_{\theta}^{n}(\mathbf{x}_{t}^{1:N},t,f(n)),\sigma_t^2 \mathbf{I} \big) \label{eq:opo_2}\\
    &\mu_{\theta}^{n}\big(\mathbf{x}_{t}^{1:N},t,f(n)\big)
    =\frac{1}{\sqrt{\alpha_t}} \bigg( \mathbf{x}_t^{n} - \frac{\beta_t}{\sqrt{1-\bar{\alpha}_t}} \epsilon_{\theta}^{n}\big(\mathbf{x}_t^{1:N},t,f(n) \big) \bigg)
    \label{eq:opo_3}
\end{align}

\subsection{Adaptation}
\label{sec:adaptation}

We adapt the pre-trained U-Net $\epsilon_{\theta}$ to the additional image $\mathbf{z}$ when supplied (later than $\mathbf{y}$) in two adaptation manners: zero-shot and optimised adaptation.

\paragraph{Zero-shot adaptation} In this setting, we apply the consistency-based gate $f$ in Eq.~(\ref{eq:condition_selection}) to determine the condition $\mathbf{c}=f(n)$, which can be $\{\mathbf{y}\}$ or $\{\mathbf{z}\}$, to be used in $\epsilon_{\theta}^{n}\big(\mathbf{x}_t^{1:N},t,f(n) \big)$ for each target view index $n \in\{1,...,N\}$. We finally apply Eq.~(\ref{eq:opo_1})-(\ref{eq:opo_3}) to synthesise views $\mathbf{x}_0^{1:N}$.

\paragraph{Optimised adaptation} In this setting, we further update $\epsilon_{\theta}$ with a denoising loss and improve cross-view consistency with a contrastive loss, given the conditions selected by $f$. We use the following denoising loss:
\begin{align}
    \mathcal{L}_{\text{denoise}} = \mathbb{E}_{t,\mathbf{x}_0^{1:N},n,\epsilon^{1:N}} \bigg\{
    \big\| \epsilon^{n} - \epsilon_{\theta}^{n}\big(\mathbf{x}_t^{1:N}, t, f(n)\big) \big\|^2_2 \bigg\}
    \label{eq:denoising_loss}    
\end{align}
where $\epsilon^{n} \sim \mathcal{N}(\mathbf{0},\mathbf{I})$ is the noise added to view $\mathbf{x}_0^{n}$.

The contrastive loss constrains the generation of views $\mathbf{x}_0^{n}$ conditioned by the newly added image $\mathbf{z}$ (i.e., $f(n)=\{\mathbf{z}\}$). In particular, we consider $\mathbf{x}_0^{n}$ an anchor and create a positive set $\mathbf{x}_0^{n+}=\{\mathbf{z}\}$ and a negative set $\mathbf{x}_0^{n-}=\{\mathbf{x}_0^{m} | m \neq n\}$. We define the loss $\ell\big(\mathbf{x}_0^{n}, \mathbf{x}_0^{n+}, \mathbf{x}_0^{n-}; \tau
\big)$ for each anchor $\mathbf{x}_0^{n}$ in the form of the NT-Xent loss in~\cite{DBLP:conf/icml/ChenK0H20} as,
\begin{align}
    \ell\big(\mathbf{x}_0^{n}, \mathbf{x}_0^{n+}, \mathbf{x}_0^{n-}; \tau
\big)=-\log \frac{\exp\big(\langle E(\mathbf{x}_0^{n}), E(\mathbf{z}) \rangle / \tau \big)}{\sum_{\mathbf{x}'\in\mathbf{x}_0^{n-}} \exp\big(\langle E(\mathbf{x}_0^{n}), E(\mathbf{x}')\rangle / \tau \big)} 
    \label{eq:pairwise_loss}
\end{align}
where $E$ is a pre-trained image encoder, $\langle \cdot,\cdot\rangle$ is the dot product, and $\tau$ is a temperature value.

We then define the contrastive loss over generated views $\mathbf{x}_0^{1:N}$ as follows,
\begin{align}
    \mathcal{L}_{\text{cont}}=\mathbb{E}_{\mathbf{x}_0^{1:N},n} \bigg\{\mathds{1}_{[f(n)=\{\mathbf{z}\}]} \ell\big(\mathbf{x}_0^{n}, \mathbf{x}_0^{n+}, \mathbf{x}_0^{n-}; \tau \big) \bigg\}
    \label{eq:contrastive_loss}
\end{align}
where $\mathds{1}_{[f(n)=\{\mathbf{z}\}]}$ indicates that $
\ell\big(\mathbf{x}_0^{n}, \mathbf{x}_0^{n+}, \mathbf{x}_0^{n-}; \tau
\big)$ is applied only to views generated with the additional image $\mathbf{z}$.

Finally, we adapt $\epsilon_{\theta}$ to the current image $\mathbf{z}$ by optimising the following loss,
\begin{align}
    \mathcal{L}_{\text{opt}} = \mathcal{L}_{\text{denoise}} + \lambda \mathcal{L}_{\text{cont}}
    \label{eq:fine_tuning_loss}
\end{align}
where $\lambda$ is a user-defined parameter.

\section{Experiments}

\subsection{Datasets}

We evaluated our method on the Google Scanned Objects (GSO) dataset~\cite{downs_google_2022}. This dataset includes 30 objects from diverse domains (e.g., household items, toys, animals). To the best of our knowledge, GSO is the only public dataset used to benchmark diffusion-based single-view 3D reconstruction. Following prior works~\cite{long_wonder3d_2024, liu_syncdreamer_2024, DBLP:conf/nips/LiLLZLLQZXLTWLG24},
we used the provided front-view image of each object as the main input and rendered another image in a random view as the additional image. We illustrate several examples of the GSO dataset in Figure~\ref{fig:qualitative_mesh_texture_GSO}. This dataset was used for quantitative evaluations and comparisons due to the availability of ground-truth data.

We also collected a small but real-world dataset of 10 objects, providing realistic data to evaluate the generalisation of our method beyond synthetic settings. Each object is provided with 2 images captured by different handheld cameras in front and auxiliary views, under various lighting and different environments. Unlike the GSO's images, our collected images contain diverse background. We present several cases of our collected images in Figure~\ref{fig:qualitative_mesh_texture_ours}. We applied the method in~\cite{Qin_2020_PR} to remove the background before passing them onto the generative models, e.g., Wonder3D (see more details in~\ref{sec:background_removal}). Note that our dataset does not include ground-truth 3D models and was therefore used for qualitative evaluations only. 

\setbox0\hbox{\tabular{@{}l}\small Single-view \\(baseline)\endtabular}

\setbox1\hbox{\tabular{@{}l}\small Sparse-\\view\endtabular}

\setbox2\hbox{\tabular{@{}l}\small Single-\\view\\Era3D\endtabular}

\setbox3\hbox{\tabular{@{}l}\small Single-\\view\\Wonder3D\endtabular}

\begin{table*}
\small
\centering
\begin{tabular}{llccccc}
\toprule
& \textbf{Method} & \textbf{CD}$\downarrow$ & \textbf{IoU}$\uparrow$ & \textbf{PSNR}$\uparrow$ & \textbf{SSIM}$\uparrow$ & \textbf{LPIPS}$\downarrow$\\
\midrule

Sparse- & SparseAGS~\cite{zhao2024sparseags} & 0.0369 & 0.4400 & 18.98 & 0.905 & 0.103 \\
view & EscherNet~\cite{DBLP:conf/cvpr/KongLLT0D24} & 0.0535 & 0.3786 & 17.32 & 0.902 & 0.127 \\
& FreeSplatter~\cite{Xu_2025_ICCV} & 0.0168 & 0.5183 & 20.28 & \textbf{0.929} & 0.073 \\
\midrule
Single-& Era3D~\cite{DBLP:conf/nips/LiLLZLLQZXLTWLG24} & 0.0171 & 0.5464 & 21.72 & 0.923 & \underline{0.067} \\
view& ASV3D (zero-shot) & \textit{0.0167} & \textit{0.5466} & \textit{21.82} & \textit{0.925} & \underline{0.067} \\
Era3D& ASV3D (optimised) & \underline{0.0164} & \underline{0.5477} & \underline{21.89} & 0.924 & \textbf{0.066} \\
\midrule
Single-& Wonder3D~\cite{long_wonder3d_2024} & 0.0218 & 0.5272 & 21.65 & 0.924 & 0.070 \\
view& ASV3D (zero-shot) & 0.0207 & 0.5355 & 21.68 & \textit{0.925} & 0.069 \\
Wonder3D& ASV3D (optimised) & \textbf{0.0120} & \textbf{0.5738} & \textbf{21.96} & \underline{0.927} & \textit{0.068} \\
\bottomrule
\end{tabular}

\caption{Comparison of ASV3D with its baselines and sparse-view 3D reconstruction methods. For each metric, the top-1, top-2, and top-3 performances are highlighted in \textbf{bold}, \underline{underline}, and \textit{italic}, respectively.}
\label{tab:eval}
\end{table*}

\subsection{Implementation Details}

We applied our method to two state-of-the-art diffusion-based single-view 3D object reconstruction baselines: Wonder3D~\cite{long_wonder3d_2024} and Era3D~\cite{DBLP:conf/nips/LiLLZLLQZXLTWLG24}.  

The zero-shot adaptation was implemented by applying the gate $f$ in Eq.~(\ref{eq:condition_selection}) to select the condition to generate each target view. For the optimised adaptation, we optimised only the layers in the U-Net that are responsible for view-consistent generation, including cross- and self-attention layers. We adopted the default setting $\tau=0.07$ in the implementation of $\ell\big(\mathbf{x}_0^{n}, \mathbf{x}_0^{n+}, \mathbf{x}_0^{n-}; \tau
\big)$ in Eq.~(\ref{eq:pairwise_loss}). We set $\lambda=0.2$ in Eq.~(\ref{eq:fine_tuning_loss}). We used $T=50$ diffusion steps. 

We used the AdamW optimiser with the learning rate set to $2.0\times10^{-5}$ and the cosine scheduler. We implemented our method in PyTorch 2.6 with xFormers accelerator. All experiments were executed on NVIDIA H200 GPUs.

\subsection{Evaluations and Comparisons}
\label{sec:evaluations_comparisons}
\paragraph{Quantitative results.} We quantitatively validate ASV3D against its baselines in both 3D reconstruction and multi-view generation in Table~\ref{tab:eval}. We measure the accuracy of 3D reconstruction by comparing reconstructed models with their ground-truth using the Chamfer Distance (CD) and the volume of Intersection-over-Union (IoU). To assess the quality of multi-view generation, we use the standard metrics of image synthesis: PSNR, SSIM, and LPIPS~\cite{DBLP:journals/tvcg/HartwigESKPPGBR25}. 

As shown, our zero-shot version applied to Wonder3D outperforms its baselines in all performance metrics. For Era3D, the zero-shot version improves the baseline in 3D reconstruction and shows advantages in 2 (out of 3) metrics for image synthesis. The optimised versions further boost the performance of both 3D reconstruction and multi-view generation, consistently shown with both baselines in all metrics. Table~\ref{tab:eval} compares our work with recent sparse-view 3D reconstruction methods despite different settings\footnote{Sparse-view reconstruction methods such as \cite{DBLP:conf/cvpr/ZhangWX000SW25, DBLP:conf/iclr/YeLXLP0P25} focus on scene reconstruction and therefore excluded in our experiments.}. Recall that our method processes the input imagery progressively rather than collectively as in sparse-view reconstruction. In addition, our method does not require pose estimation. In all, our optimised adaptation of Wonder3D achieves state-of-the-art performance in both 3D reconstruction and multi-view generation.

\begin{figure}[ht]
\centering
\includegraphics[width=1.0\linewidth]{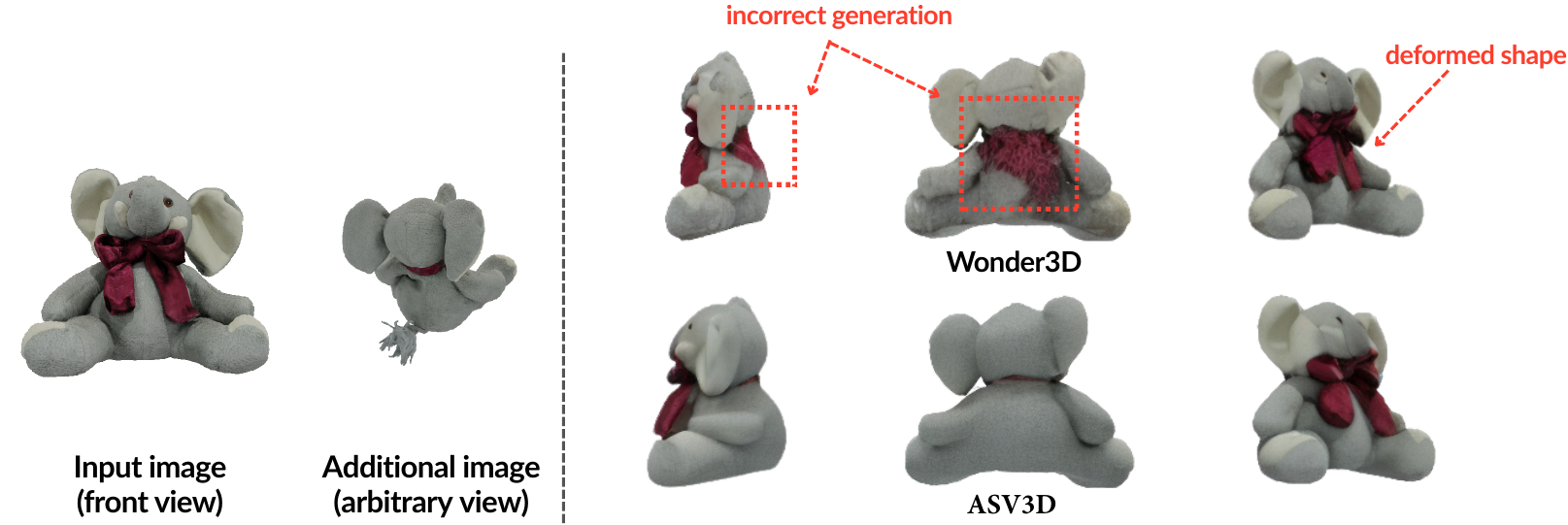} \\
\text{(a)}


\includegraphics[width=1.0\linewidth]{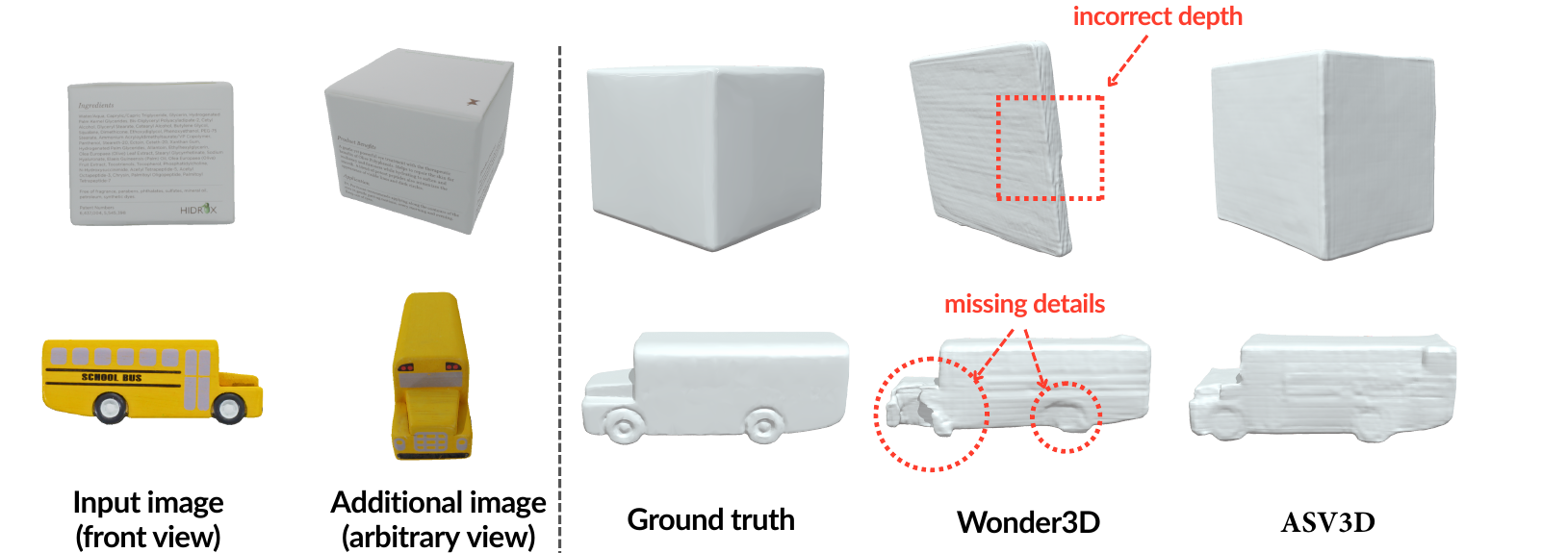} \\
\text{(b)}

\caption{\textbf{Qualitative comparison of ASV3D with Wonder3D baseline} in 3D reconstruction (a) and in multi-view image generation (b). Artifacts in the reconstruction and generation results are highlighted.}
\label{fig:qualitative_comparison}
\end{figure}

\paragraph{Qualitative results.} 

Figure~\ref{fig:qualitative_comparison} shows how our optimised adaptation improves Wonder3D. As presented in Figure~\ref{fig:qualitative_comparison}(a), the 3D reconstruction result of Wonder3D in the 1\textit{st} example reveals shape deformation and incorrect depth reconstruction. The reconstructed model by Wonder3D in the 2\textit{nd} example shows missing details. In contrast, ASV3D consistently recovers more complete geometry and preserves fine structures, e.g., rims, planar faces, and thickness.

The multi-view generation results show the cross-view inconsistency produced by Wonder3D. For instance, the additional image in Figure~\ref{fig:qualitative_comparison}(a) shows that the scarf is not visible from the back view. However, Wonder3D generates it in both side and back views. This example also highlights the usefulness of the additional image. Figure~\ref{fig:qualitative_comparison}(b) shows that our method yields noticeably more coherent appearance, particularly on the back/occluded sides that are under-constrained from the front view alone. The auxiliary view provides complementary cues that restore texture and suppress the flattening artifacts typical of single-view reconstructions. Finally, thanks to contrastive learning, our optimised adaptation produces smoother viewpoint transitions, improving multi-view consistency without overfitting to specific poses. 

\begin{figure}
\centering
\includegraphics[width=1.0\linewidth]{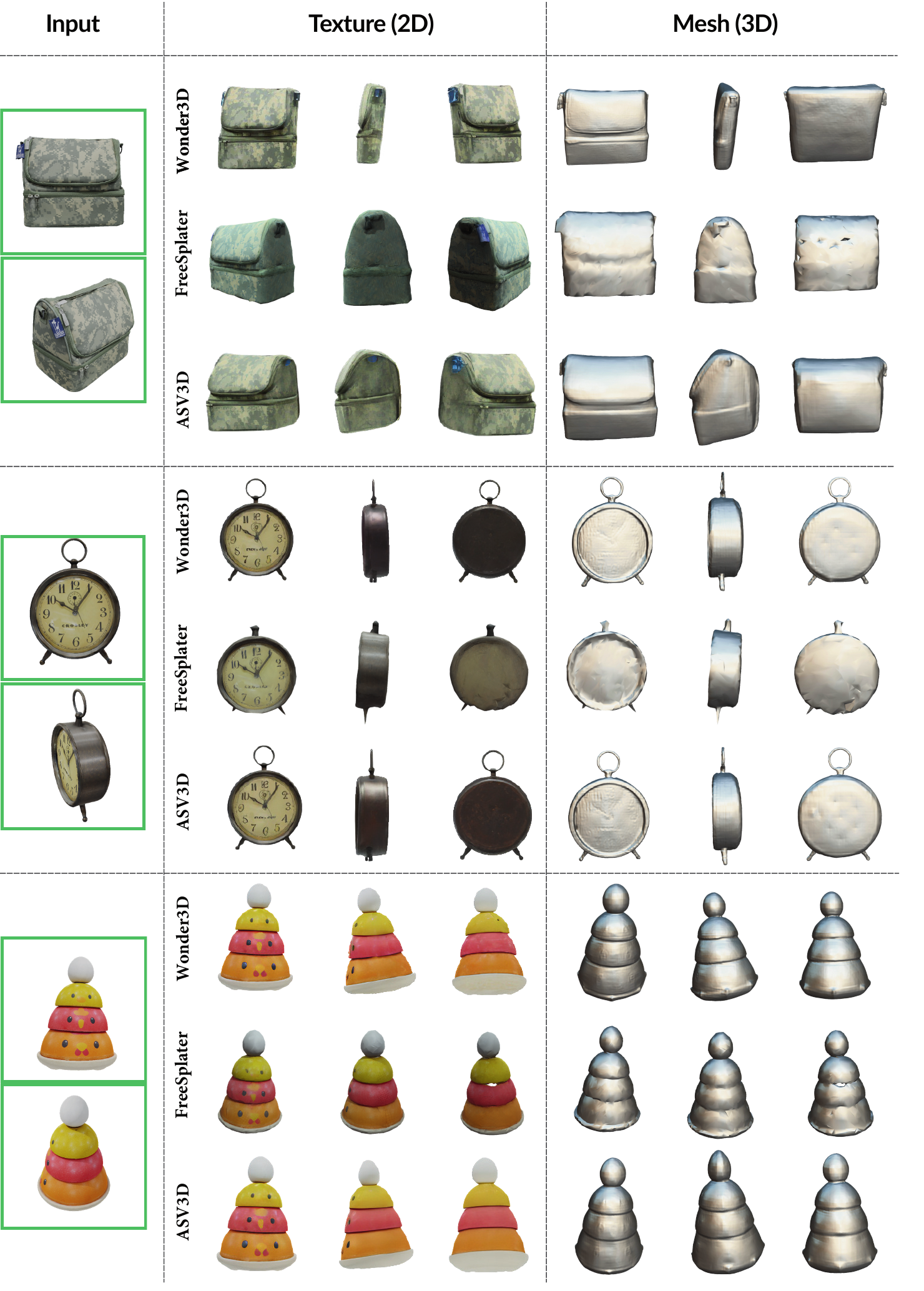}
\caption{\textbf{Qualitative results of ASV3D (optimised version), Wonder3D baseline, and FreeSplatter on GSO objects}. Left: Input image (top) and additional image (bottom). Middle: Multi-view images. Right: 3D reconstruction results.}
\label{fig:qualitative_mesh_texture_GSO}
\end{figure}

\begin{figure}
\centering
\includegraphics[width=1.0\linewidth]{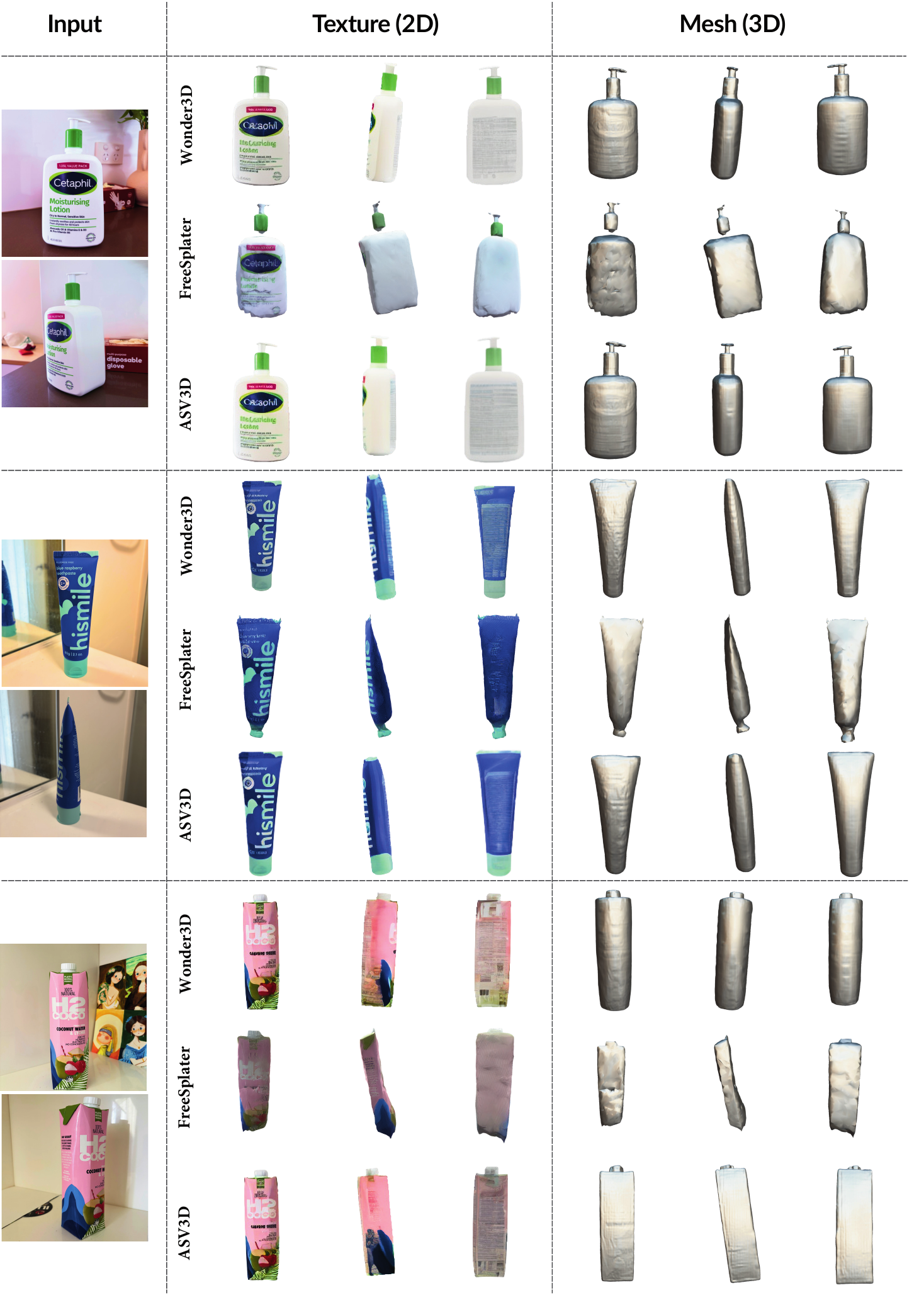}
\caption{\textbf{Qualitative results of ASV3D (optimised version), Wonder3D baseline, and FreeSplatter on real-world objects}. Left: Input image (top) and additional image (bottom). Middle: Multi-view images. Right: 3D reconstruction results.}
\label{fig:mv_real}
\end{figure}

Figure~\ref{fig:qualitative_mesh_texture_GSO} and Figure~\ref{fig:mv_real} compare our ASV3D (optimised adaptation) with the Wonder3D baseline~\cite{long_wonder3d_2024} and the sparse-view 3D object reconstruction method FreeSplatter~\cite{Xu_2025_ICCV} on GSO and our collected real-world objects, respectively. These competing methods are chosen as they show state-of-the-art performance within their field (i.e., FreeSplatter) or when adapted with our method (i.e., Wonder3D), as shown in Table~\ref{tab:eval}. For each example, we provide a primary front-view image and an auxiliary image captured from an arbitrary viewpoint with a different camera. 

\begin{figure}
\centering
\includegraphics[width=1.0\linewidth]{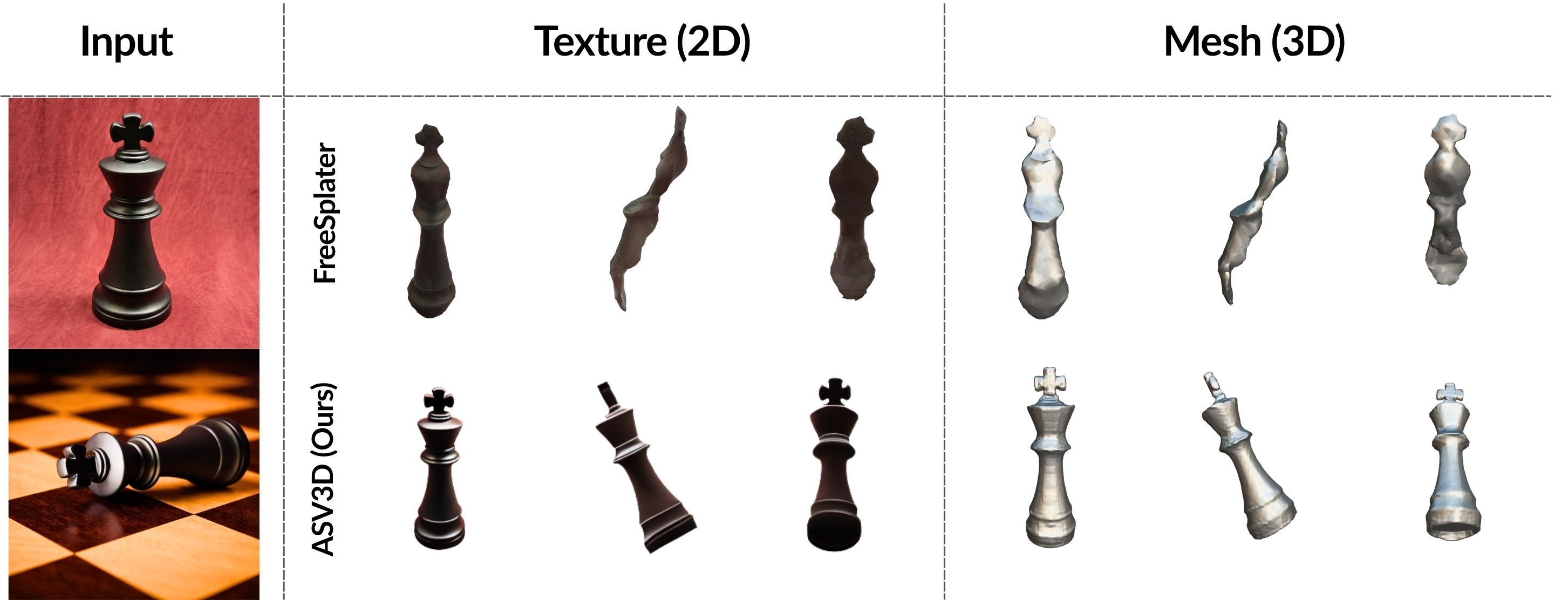}
\caption{\textbf{Qualitative comparison of ASV3D (optimised version) with FreeSplatter under an extreme pose gap.} Left: Input image (top) and additional image (bottom). Middle: Multi-view images. Right: 3D reconstruction results.}
\label{fig:mv_real_2}
\end{figure}

\begin{figure}
\centering
\includegraphics[width=1.0\linewidth]{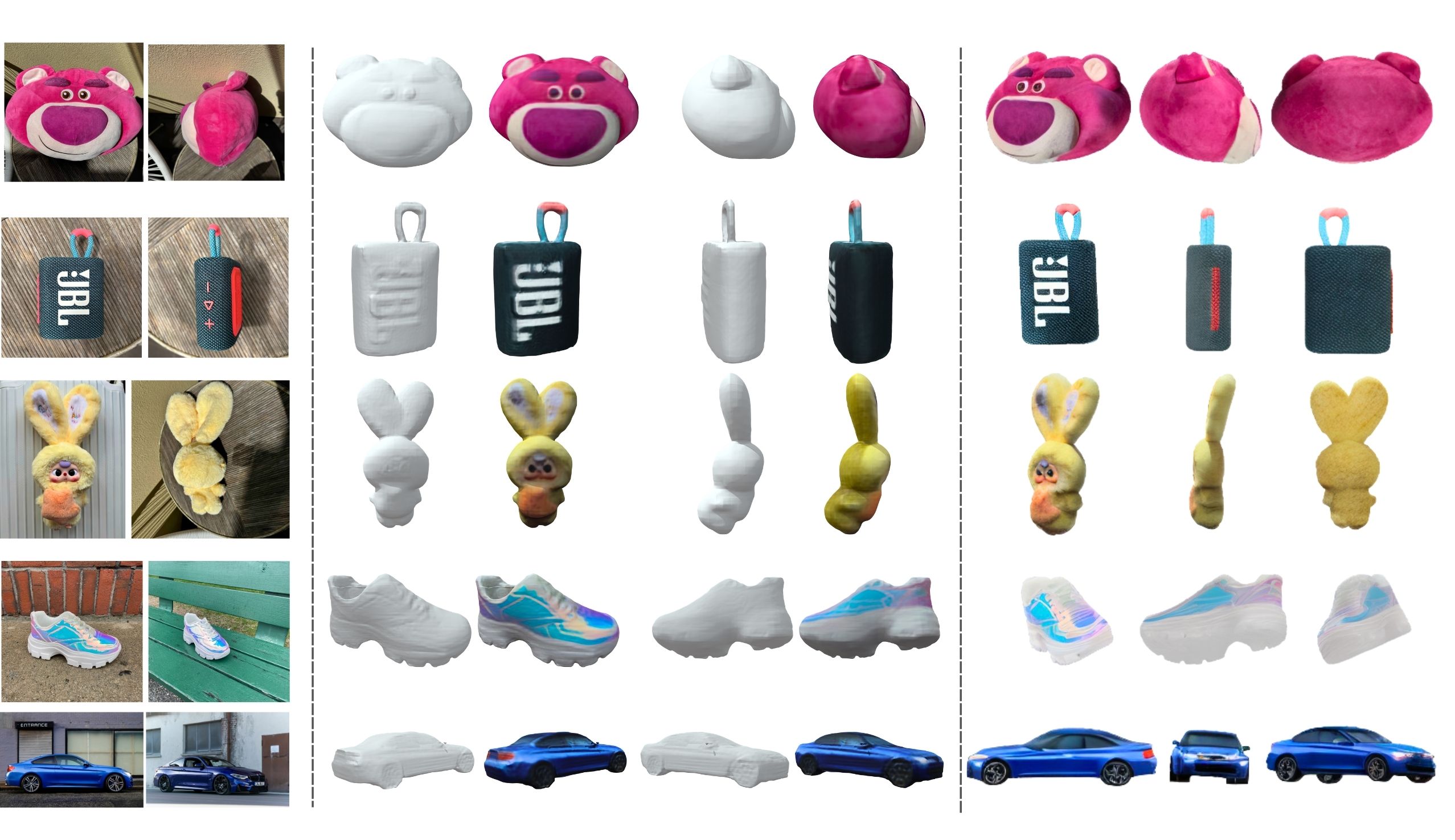}
\caption{\textbf{Qualitative results of ASV3D (optimised version) on real-world objects}. (Left) Input front and auxiliary images. 
(Middle) Reconstructed mesh and rendered texture. 
(Right) Generated views. As shown, ASV3D yields consistent geometry, accurate texture, and stable multi-view appearance across diverse object categories and scales.}
\label{fig:qualitative_mesh_texture_ours}
\end{figure}

We found that FreeSplatter works well on GSO objects, sharing the same domain with its training data, but fails to generalise to real-world objects when the primary and additional images are captured at different distances from the object, and especially in pose gap situations (see Figure~\ref{fig:mv_real_2}). In contrast, ASV3D maintains stable geometry, faithful appearance, and robust spatial coherence under varied object poses and lighting conditions. We provide additional qualitative results of ASV3D on the collected real-world objects in Figure~\ref{fig:qualitative_mesh_texture_ours}. 

\paragraph{Computational analysis.} Recall that our ASV3D first applies the gate $f$ to determine the condition for each target view. Our zero-shot variant then generates multiple views $\mathbf{x}_0^{1:N}$ based on the formed conditions. The optimised variant updates the model with current data. We provide details of the computational cost of our pipeline applied to Wonder3D in Table~\ref{tab:computation}, where Wonder3D corresponds to columns 3 and 4, the zero-shot version corresponds to columns 1, 3, 4, and the optimised version uses all columns. Overall, our zero-shot and optimised versions complete the same task in approximately {3 min 51 s and 7 min}, respectively.

\begin{table}[ht]
\small
\centering
\setlength{\tabcolsep}{5pt}
\begin{tabular}{cccc}
\toprule
\textbf{Apply} $f$ & \textbf{Optimised adapt.} & \textbf{Multi-view syn.} & \textbf{3D reconst.} \\
\midrule
1 s & 3 min 9 s & 2 min 20 s & 1 min 30 s \\
\bottomrule \\
\end{tabular}

\caption{Computational cost of ASV3D.}
\label{tab:computation}
\end{table}

\subsection{Ablation Study}
\label{sec:ablation}

\begin{table}
\small
\centering

\setlength{\tabcolsep}{5pt}
\renewcommand{\arraystretch}{1.05}
\begin{tabular}{lccccc}
\toprule
\textbf{Setting} & \textbf{CD}$\downarrow$ & \textbf{IoU}$\uparrow$ & \textbf{PSNR}$\uparrow$ & \textbf{SSIM}$\uparrow$ & \textbf{LPIPS}$\downarrow$ \\
\midrule
Multi-cond. & 0.0228 & 0.5198 & 21.66 & 0.9247 & 0.070 \\
Our gate $f$ & \textbf{0.0207} & \textbf{0.5355} & \textbf{21.68} & \textbf{0.9250} & \textbf{0.069} \\
\bottomrule \\
\end{tabular}

\caption{Comparison of our ASV3D image conditioning and multi-image conditioning in our zero-shot variant on GSO objects. Best performance for each metric is highlighted.}

\label{tab:naive_zero_shot}

\end{table}

\paragraph{When does an extra view help?} As argued earlier, an extra view is not always helpful to multi-view synthesis. Our proposed consistency-based gate $f$ can be used to decide if the additional image $\mathbf{z}$ or the primary input image $\mathbf{y}$ should be used to condition the generation of a target view. Diffusion-based single-view 3D reconstruction methods (e.g., Wonder3D, Era3D) always use the primary input image $\mathbf{y}$ as the only condition to generate all target views. The multi-conditioning approach~\cite{DBLP:conf/cvpr/YangHWGZZ0L24}, on the other hand, concatenates $\mathbf{y}$ and $\mathbf{z}$ to form a new condition which is used in all view generation. Our approach chooses between $\mathbf{y}$ and $\mathbf{z}$ based on their consistency in generation. To prove the effectiveness of our consistency-based gating mechanism, we compare it with the multi-conditioning approach~\cite{DBLP:conf/cvpr/YangHWGZZ0L24}. We ran this experiment with Wonder3D since it can be extended to adapt with multi-conditioning. 

Table~\ref{tab:naive_zero_shot} shows that multi-conditioning made up of all input images is not effective, as some input images may be irrelevant, misleading the generation process. In contrast, our approach finds the optimal conditions for all target view generations. This result also indicates that the improvement achieved by our method is due to the conditioning strategy rather than to the sole use of auxiliary imagery. In addition, the consistency-based gate can tell us when the extra image $\mathbf{z}$ is useful. 

\paragraph{Consistency-based gating} The consistency-based gate $f$ presented in Eq.~(\ref{eq:condition_selection}) addresses the central difficulty of using an auxiliary image to condition the generation of target views without requiring camera pose estimation. Specifically, the gate $f$ determines which input (primary image or additional image) provides reliable evidence for each target view generation. Inspired by the consistency models~\cite{DBLP:conf/icml/SongD0S23, DBLP:conf/nips/DarasDDD23}, we design $f$ based on the denoising convergence of the U-Net $\epsilon_{\theta}$. Here, we validate our proposed consistency-based gating in two different settings and compare it with another approach. 

As presented, we calculate the consistency of $\epsilon_{\theta}$ in the late half diffusion steps $\mathcal{T}=\{\frac{T}{2},...,1\}$ because we found that the early denoising results are not stable. To validate this finding, we compare our half trajectory setting with the full trajectory setting, i.e., $\mathcal{T}=\{T,...,1\}$. In addition, we compare our consistency-based metric with the sampling trajectory divergence (STD) used in trajectory distillation models~\cite{DBLP:conf/iccv/LuoHSCT25}. The STD of a diffusion model can be estimated as the mean squared error between the outputs generated with the exact same seed but different step counts. In this experiment, we set the step counts for $\epsilon_{\theta}$ to 50 (fine trajectory) and 25 (course trajectory).

We evaluate each gating strategy by measuring the Spearman correlation between the consistency score of the gating strategy for a condition (i.e., the input $\mathbf{y}$ or the additional image $\mathbf{z}$) and the similarity between the respective generation result and its ground truth. Specifically, for each target view $n$, let $g(\mathbf{c})$ be the consistency score of the generative model with respect to a condition $\mathbf{c} \in \mathbf{C}=\{\mathbf{y},\mathbf{z}\}$ in the  prediction of view $n$. For our consistency-based gate, $g(\mathbf{c})=\mathbb{E}_{t\in\mathcal{T}}\big\{\|\mu_{\theta}^{n}\big(\mathbf{x}_{t}^{1:N},t,\mathbf{c}\big)-\hat{\mu}\big\|_2^2\big\}$ where $\hat{\mu}=\mathbb{E}_{t\in\mathcal{T}}\big\{\mu_{\theta}^{n}\big(\mathbf{x}_{t}^{1:N},t,\mathbf{c}\big)\big\}$ and $\mathcal{T}=\{\frac{T}{2},...,1\}$. For the full trajectory setting, the same formula for $g(\mathbf{c})$ is used but for $\mathcal{T}=\{T,...,1\}$. For STD~\cite{DBLP:conf/iccv/LuoHSCT25}, we define $g(\mathbf{c})=\big\|\mu_{\theta, \text{fine}}^{n}\big(\mathbf{x}_{1}^{1:N},1,\mathbf{c}\big)-\mu_{\theta, \text{course}}^{n}\big(\mathbf{x}_{1}^{1:N},1,\mathbf{c}\big)\big\|_2^2$, where $\mu_{\theta, \text{fine}}$ and $\mu_{\theta, \text{course}}$ are estimates with step counts set to 50 (fine trajectory) and 25 (course trajectory), respectively.

Let $s(\mathbf{c})$ be the similarity between the predicted result in view $n$ (i.e., $\mu_{\theta}^{n}\big(\mathbf{x}_{1}^{1:N},1,\mathbf{c}\big)$) and its ground truth. We measure the Spearman correlation coefficient between $\{g(\mathbf{c})\}_{\mathbf{c}\in\mathbf{C}}$ and $\{s(\mathbf{c})\}_{\mathbf{c}\in\mathbf{C}}$. The Spearman correlation coefficient reflects the reliability of the gating mechanism. Intuitively, higher generation consistency implies better condition selection, leading to better generation.

\begin{figure*}[ht]
\centering
\includegraphics[width=0.8\linewidth]{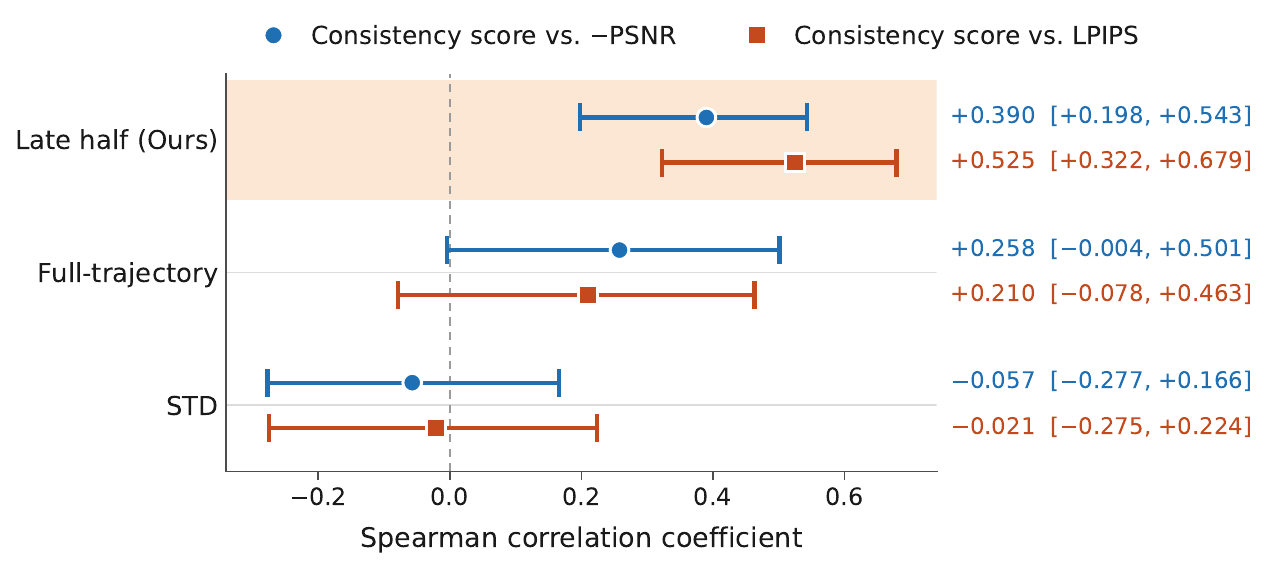}
\caption{\textbf{Validation of consistency-based gating.} Among the tested measurements, our proposed consistency-based gate with late half diffusion steps shows the strongest observed pooled association with poorer condition quality for both negative PSNR and LPIPS. Error bars denote object-cluster bootstrap 95\% confidence intervals.}
\label{fig:proxy_pooled_validation}
\end{figure*}

We used PSNR and LPIPS as image similarity metrics. We report the results of this experiment on GSO objects in Figure~\ref{fig:proxy_pooled_validation}. As shown in the results, our proposed consistency-based gate (with the late half setting) achieves both highest correlation with image generation quality and the lowest measure variance across image similarity metrics. Note that because lower negative PSNR and higher LPIPS indicate poorer synthesis, positive correlations mean that a larger trajectory dispersion is associated with a less reliable condition.

\paragraph{Optimised adaptation loss.} We studied the impact of the loss functions: $\mathcal{L}_{\text{denoise}}$ in Eq.~(\ref{eq:denoising_loss}) and $\mathcal{L}_{\text{cont}}$ in Eq.~(\ref{eq:contrastive_loss}). Specifically, we conducted the following experiments: (i) optimisation with denoising loss $\mathcal{L}_{\text{denoise}}$ only (i.e., $\mathcal{L}_{\text{opt}}=\mathcal{L}_{\text{denoise}}$), (ii) optimisation with contrastive loss $\mathcal{L}_{\text{cont}}$ only (i.e., $\mathcal{L}_{\text{opt}}=\mathcal{L}_{\text{cont}}$), and (iii) using both loss functions as in Eq.~(\ref{eq:fine_tuning_loss}). For scenario (iii), we also experimented with $\lambda=0.1$ and $\lambda=0.2$. We report the results of this study with Wonder3D baseline in Table~\ref{tab:ablation_modules}. The experimental results confirm the role of each loss function and their combination. As shown, the best performance, for both 3D reconstruction and multi-view generation, is achieved when both $\mathcal{L}_{\text{denoise}}$ and $\mathcal{L}_{\text{cont}}$ are combined with $\lambda=0.2$.

\begin{table}[ht]
\small
\centering

\setlength{\tabcolsep}{5pt}
\renewcommand{\arraystretch}{1.05}
\begin{tabular}{lccccc}
\toprule
\textbf{Setting} & \textbf{CD}$\downarrow$ & \textbf{IoU}$\uparrow$ & \textbf{PSNR}$\uparrow$ & \textbf{SSIM}$\uparrow$ & \textbf{LPIPS}$\downarrow$ \\
\midrule
$\mathcal{L}_{\text{denoise}}$                 & 0.0155 & 0.5637 & 21.89 & 0.926 & 0.068 \\
$\mathcal{L}_{\text{cont}}$               & 0.0184 & 0.5495 & 21.81 & 0.925 & 0.069 \\
$\mathcal{L}_{\text{opt}}$ ($\lambda=0.1$)          & 0.0121 & 0.5705 & 21.96 & 0.927 & 0.068 \\
$\mathcal{L}_{\text{opt}}$  ($\lambda=0.2$)   
& \textbf{0.0120} & \textbf{0.5738} & \textbf{21.96} & \textbf{0.927} & \textbf{0.068} \\
\bottomrule \\
\end{tabular}
\caption{Impact of the loss functions used in our optimised adaptation. Best performances are highlighted.}
\label{tab:ablation_modules}
\end{table}

\subsection{User Study}
\label{sec:user_study}

We conducted a user study with 32 participants. The study aims to compare ASV3D (optimised version) with its Wonder3D baseline in two tasks: 3D reconstruction (task 1) and multi-view image generation (task 2). Each task includes the results by ASV3D and Wonder3D on 10 objects: 5 objects from GSO and 5 objects from our real-world dataset. All participants were required to provide their ratings for the quality of the multi-view generation and 3D reconstruction results (e.g., alignment of a reconstructed mesh with its reference images for task 1, and photo-realism and consistency of generated views for task 2). Ratings were defined on a 5-point Likert scale (1 = not aligned/realistic, 5 = extremely aligned/realistic). Details of the user study are presented in~\ref{sec:user_study_details}.

\begin{figure}[ht]
\centering
\includegraphics[width=1.0\linewidth]{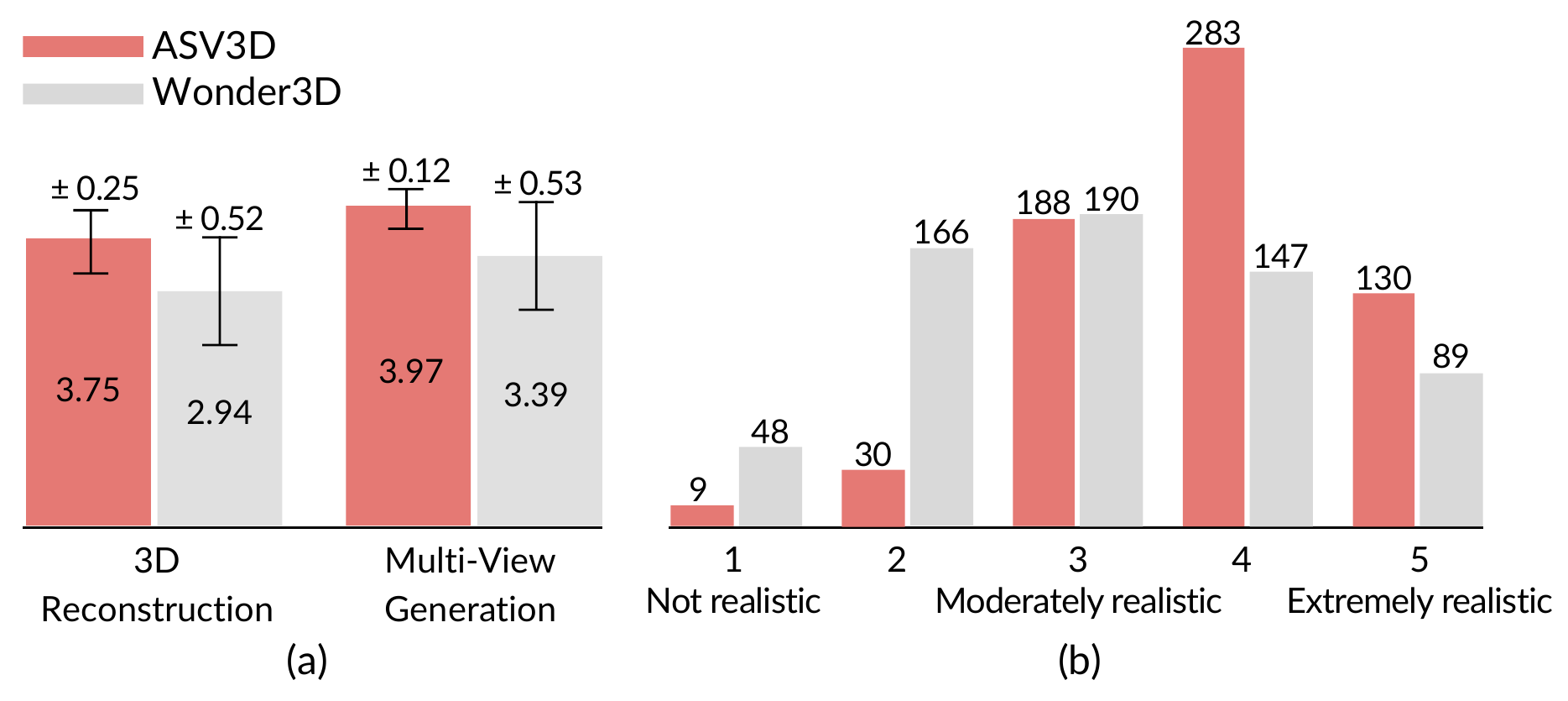}
\caption{\textbf{User study results}.
(a) Mean score and standard deviation of ratings.
(b) Rating distributions (summarised from both tasks).
}

\label{fig:user_study_results}
\end{figure}

Figure~\ref{fig:user_study_results} summarises user preference. As shown, ASV3D is consistently favoured to Wonder3D in both tasks, evident by higher mean scores and lower standard deviations. 

\section{Conclusion}

This paper introduces ASV3D, a method for adapting diffusion-based single-view 3D reconstruction to test-time data with a primary input image and an additional image, both from different cameras. We propose handling such input in a diffusion-based single-view 3D reconstruction pipeline using a consistency-based gating mechanism, which effectively determines the condition for the diffusion model. Based on this gating approach, we devise two adaptation strategies: zero-shot and optimised adaptation. The zero-shot adaptation directly feeds the additional imagery to a pre-trained diffusion-based single-view 3D reconstruction model using the consistency-based gate. The optimised adaptation further enhances cross-view consistency via contrastive learning, improving both multi-view rendering and 3D reconstruction. Extensive experiments with state-of-the-art diffusion-based single-view 3D reconstruction baselines on benchmark and real-world datasets confirm the robustness of our method and its superiority over the baselines and other approaches. Our method has the potential for a new capacity: novel-view input acquisition, i.e., the method can suggest an auxiliary view to be acquired to improve the reconstruction of a current or pre-reconstructed object. We consider this direction for our future work.

\appendix

\section{Background Handling}
\label{sec:background_removal}

As in the standard 3D object reconstruction protocol, an object's geometry and texture can be reconstructed from the foreground presented in the object's images. The reconstruction can be done effectively with the object's images containing no background. Our real-world dataset deliberately presents more challenging yet realistic scenarios in which the two images are captured with different cameras, under different illumination conditions, and against different backgrounds (see Figure~\ref{fig:mv_real} and Figure~\ref{fig:qualitative_mesh_texture_ours}). 

\begin{figure}
\centering
\includegraphics[width=0.9\linewidth]{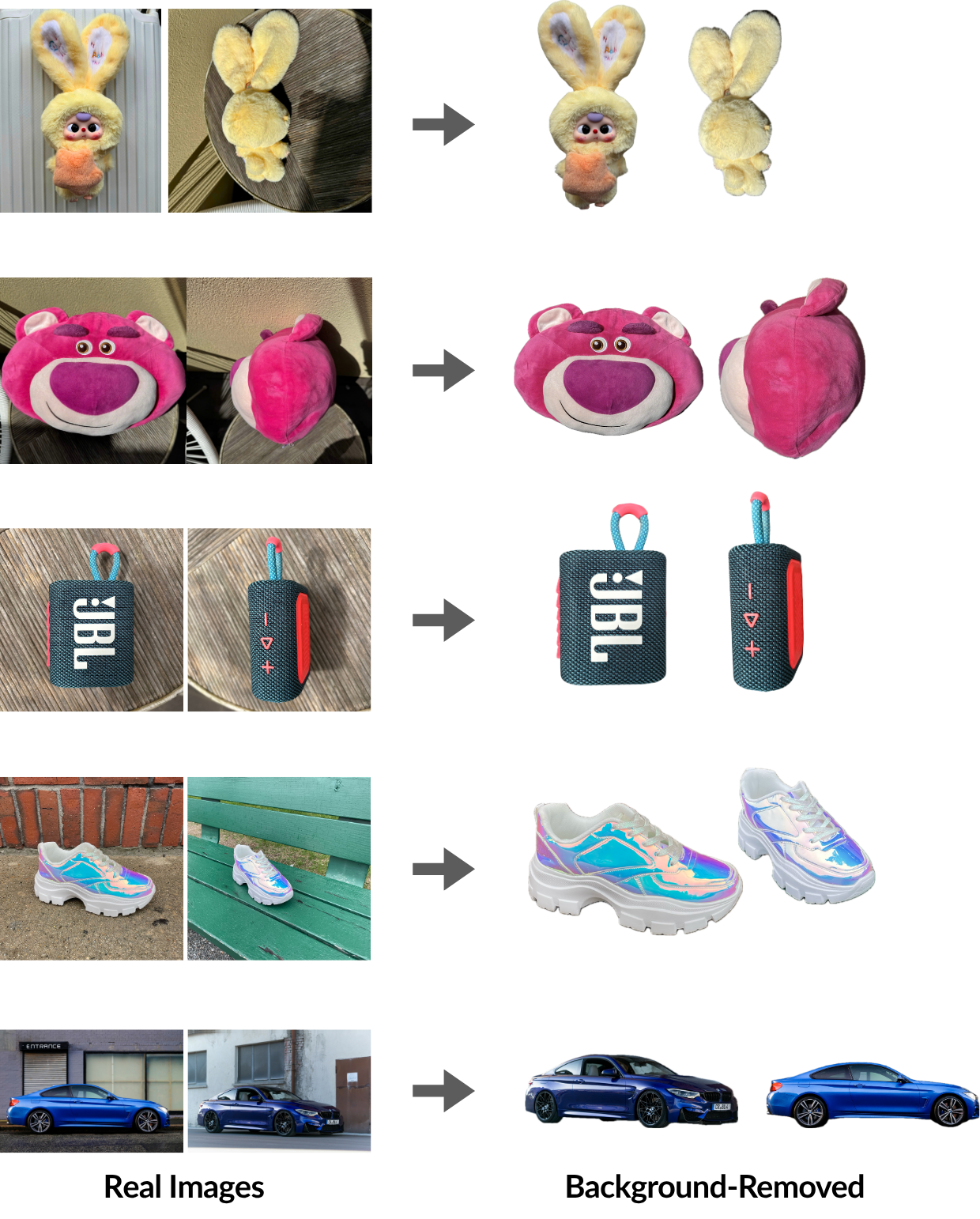}
\caption{\textbf{Background removal on real-world inputs.} Original images from our collected dataset (left) and foreground segmentation results produced by~\cite{Qin_2020_PR} (right).}
\label{fig:real_cutouts}
\end{figure}

Following the standard reconstruction fashion, we remove the background from an input image prior to passing the image to the reconstruction pipeline. In our work, we applied the method in~\cite{Qin_2020_PR} that uses a U-net architecture to segment the foreground object from an input image. We illustrate several foreground segmentation results from our collected dataset in Figure~\ref{fig:real_cutouts}. We found that this approach works well under various illumination conditions and clutter. Despite the imperfections observed, the segmented foreground regions maintain enough detail for our method to perform multi-view generation and then 3D reconstruction (see Figure~\ref{fig:mv_real} and Figure~\ref{fig:qualitative_mesh_texture_ours}). 

\begin{figure}
\centering
\includegraphics[width=0.9\linewidth]{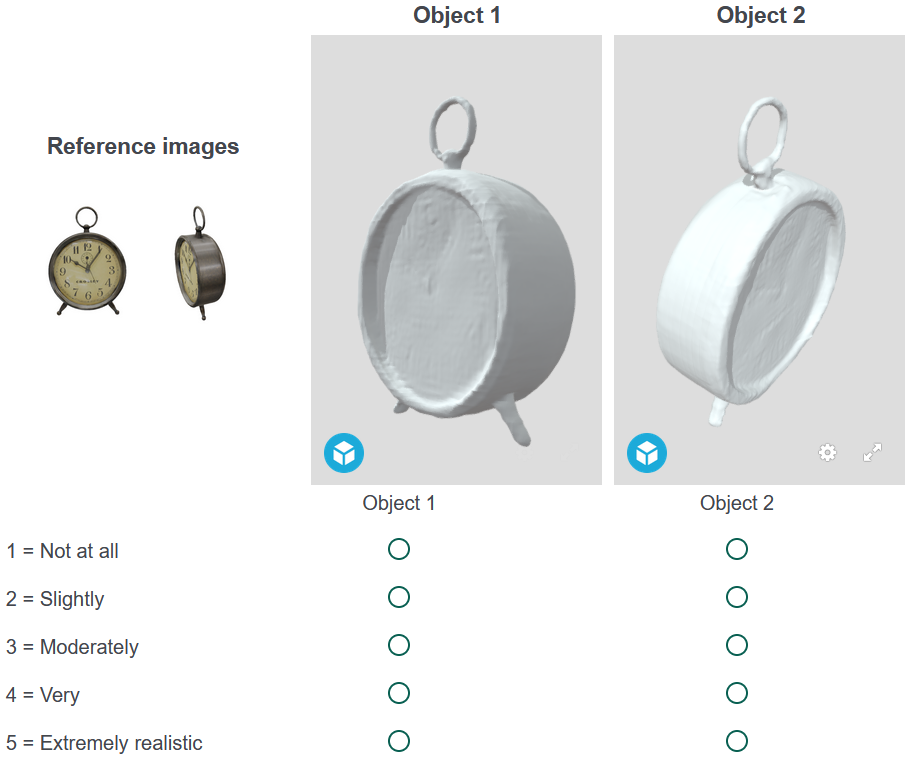}\\
\text{(a)}\\
\includegraphics[width=1.0\linewidth]{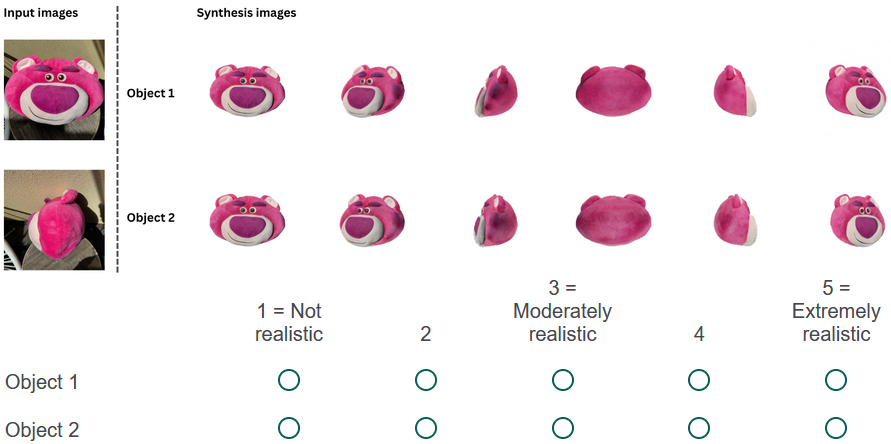}\\
\text{(b)}
\caption{\textbf{Example user-study interfaces} for evaluating (a) 3D reconstruction and (b) multi-view image generation. Method identities and display positions were randomised during the study.}
\label{fig:survey}
\end{figure}

\section{User Study Design}
\label{sec:user_study_details}

\subsection{Materials \& Participants}

Our study aims to understand the quality of our ASV3D (optimised version) and its baseline (Wonder3D) in 3D reconstruction and multi-view synthesis from user perspective. We used the results of our ASV3D and Wonder3D on 10 objects: 5 objects (from the GSO dataset) and 5 objects (from our real-world dataset) were randomly selected. For each object, a pair of an input image (in the front view) and an additional image (in an arbitrary view) is given. The sample size for the entire survey was set at 10, following established Human-Computer Interaction (HCI) guidelines to maintain the duration of the user study in 10 minutes to ensure high response quality and minimal fatigue of the participants~\cite{galesic2009effects}. Interactive 3D viewers and reference images were embedded directly within the survey with consistent lighting and camera settings.

We recruited 32 participants (18+ years) with normal computer skills (ability to interact with computers and 3D viewer software to examine the quality of 3D reconstruction) and with common understanding of 3D space.  

\subsection{Procedure}

Each participant evaluated the quality of the results made by ASV3D and Wonder3D from each input sample across two tasks: 3D reconstruction (task 1) and multi-view image generation (task 2) as follows.
\begin{itemize}
    \item \textbf{Task 1}. The participants viewed two reference images along with two 3D reconstructions (\textbf{ASV3D} vs.\ \textbf{Wonder3D}) displayed in an interactive viewer. They could freely rotate, zoom, and inspect each model. The participants rated the geometric realism of each reconstruction:
    \begin{quote}
    \emph{“How realistic does this 3D reconstruction appear compared to the two reference images?”}
    \end{quote}
    We used a 5-point Likert scale for the ratings, where 1 = not realistic, 5 = extremely realistic.

    \item \textbf{Task 2}. The participants saw two reference images and two rows of synthesized views (rendered from the reconstructed meshes). They rated the photo-realism and cross-view consistency of each result:
    \begin{quote}
        \emph{“How realistic are the textures and materials in the synthesized views?”}
    \end{quote}
    We used the same five-point Likert scale.
\end{itemize}

For both tasks, the participants also selected a forced-choice preference  
(\textbf{ASV3D} / \textbf{Wonder3D} / \textbf{Tie}). To avoid systematic bias, the order of the samples and the placement of the results (one from ASV3D and one from Wonder3D) were randomised. In addition, all stimuli were standardised to ensure fair comparison: identical render settings and equal viewpoint spacing for both ASV3D and Wonder3D. We show example questions in our survey in Figure~\ref{fig:survey}. 

\subsection{Data Management and Ethics}

The study was administered online through a secure survey platform. No personally identifiable information was requested and responses were recorded anonymously. The survey data were stored on encrypted institutional servers with restricted access and will be retained for up to five years.

\small
\bibliographystyle{elsarticle-num}
\bibliography{ref}

@inproceedings{DBLP:conf/iccv/LuoHSCT25,
  author       = {Yihong Luo and
                  Tianyang Hu and
                  Jiacheng Sun and
                  Yujun Cai and
                  Jing Tang},
  title        = {Learning Few-Step Diffusion Models by Trajectory Distribution Matching},
  booktitle    = {{IEEE/CVF} International Conference on Computer Vision},
  pages        = {17719--17728},
  year         = {2025}
}

@inproceedings{DBLP:conf/cvpr/KongLLT0D24,
  author       = {Xin Kong and
                  Shikun Liu and
                  Xiaoyang Lyu and
                  Marwan Taher and
                  Xiaojuan Qi and
                  Andrew J. Davison},
  title        = {{EscherNet}: {A} Generative Model for Scalable View Synthesis},
  booktitle    = {{IEEE/CVF} Conference on Computer Vision and Pattern Recognition},
  pages        = {9503--9513},
  year         = {2024}
}

@inproceedings{DBLP:conf/icml/SongD0S23,
  author       = {Yang Song and
                  Prafulla Dhariwal and
                  Mark Chen and
                  Ilya Sutskever},
  title        = {Consistency Models},
  booktitle    = {International Conference on Machine Learning},
  pages        = {32211--32252},
  year         = {2023}
}

@inproceedings{DBLP:conf/cvpr/LeungHV22,
  author       = {Brandon Leung and
                  Chih{-}Hui Ho and
                  Nuno Vasconcelos},
  title        = {Black-Box Test-Time Shape {REFINEment} for Single View {3D} Reconstruction},
  booktitle    = {{IEEE/CVF} Conference on Computer Vision and Pattern Recognition Workshops},
  pages        = {4079--4089},
  year         = {2022}
}

@inproceedings{DBLP:conf/bmvc/Yu-JiHYSHO24,
  author       = {Kim Yu{-}Ji and
                  Hyunwoo Ha and
                  Kim Youwang and
                  Jaeheung Surh and
                  Hyowon Ha and
                  Tae{-}Hyun Oh},
  title        = {{MeTTA}: Single-View to {3D} Textured Mesh Reconstruction with Test-Time
                  Adaptation},
  booktitle    = {British Machine Vision Conference},
  year         = {2024}
}

@inproceedings{DBLP:conf/nips/YuanSWYW25,
  author       = {Yuheng Yuan and
                  Qiuhong Shen and
                  Shizun Wang and
                  Xingyi Yang and
                  Xinchao Wang},
  title        = {{Test3R}: Learning to Reconstruct {3D} at Test Time},
  booktitle    = {Advances in Neural Information Processing Systems},
  year         = {2025}
}

@inproceedings{DBLP:journals/corr/abs-2509-26645,
  author       = {Xingyu Chen and
                  Yue Chen and
                  Yuliang Xiu and
                  Andreas Geiger and
                  Anpei Chen},
  title        = {{TTT3R:} {3D} Reconstruction as Test-Time Training},
  booktitle    = {International Conference on Learning Representations},
  year         = {2026}
}

@inproceedings{DBLP:conf/cvpr/ZhangWX000SW25,
  author       = {Shangzhan Zhang and
                  Jianyuan Wang and
                  Yinghao Xu and
                  Nan Xue and
                  Christian Rupprecht and
                  Xiaowei Zhou and
                  Yujun Shen and
                  Gordon Wetzstein},
  title        = {{FLARE:} Feed-forward Geometry, Appearance and Camera Estimation from
                  Uncalibrated Sparse Views},
  booktitle    = {{IEEE/CVF} Conference on Computer Vision and Pattern Recognition},
  pages        = {21936--21947},
  year         = {2025}
}

@inproceedings{DBLP:conf/iclr/YeLXLP0P25,
  author       = {Botao Ye and
                  Sifei Liu and
                  Haofei Xu and
                  Xueting Li and
                  Marc Pollefeys and
                  Ming{-}Hsuan Yang and
                  Songyou Peng},
  title        = {No Pose, No Problem: Surprisingly Simple {3D} {Gaussian} Splats from Sparse
                  Unposed Images},
  booktitle    = {International Conference on Learning Representations},
  year         = {2025}
}

@inproceedings{DBLP:conf/cvpr/Richter018,
  author       = {Stephan R. Richter and
                  Stefan Roth},
  title        = {Matryoshka Networks: Predicting {3D} Geometry via Nested Shape Layers},
  booktitle    = {{IEEE/CVF} Conference on Computer Vision and Pattern Recognition},
  pages        = {1936--1944},
  year         = {2018}
}

@inproceedings{DBLP:conf/cvpr/GroueixFKRA18,
  author       = {Thibault Groueix and
                  Matthew Fisher and
                  Vladimir G. Kim and
                  Bryan C. Russell and
                  Mathieu Aubry},
  title        = {A Papier-M{\^{a}}ch{\'{e}} Approach to Learning {3D} Surface Generation},
  booktitle    = {{IEEE/CVF} Conference on Computer Vision and Pattern Recognition},
  pages        = {216--224},
  year         = {2018}
}

@inproceedings{DBLP:conf/iccv/TatarchenkoDB17,
  author       = {Maxim Tatarchenko and
                  Alexey Dosovitskiy and
                  Thomas Brox},
  title        = {Octree Generating Networks: Efficient Convolutional Architectures
                  for High-resolution {3D} Outputs},
  booktitle    = {{IEEE/CVF} International Conference on Computer Vision},
  pages        = {2107--2115},
  year         = {2017}
}

@inproceedings{DBLP:conf/nips/0001ZXFT16,
  author       = {Jiajun Wu and
                  Chengkai Zhang and
                  Tianfan Xue and
                  Bill Freeman and
                  Josh Tenenbaum},
  title        = {Learning a Probabilistic Latent Space of Object Shapes via {3D} Generative-Adversarial
                  Modeling},
  booktitle    = {Advances in Neural Information Processing Systems},
  year         = {2016}
}

@inproceedings{DBLP:conf/eccv/GirdharFRG16,
  author       = {Rohit Girdhar and
                  David F. Fouhey and
                  Mikel Rodriguez and
                  Abhinav Gupta},
  title        = {Learning a Predictable and Generative Vector Representation for Objects},
  booktitle    = {European Conference Computer Vision},
  pages        = {484--499},
  year         = {2016}
}

@inproceedings{DBLP:conf/eccv/ChoyXGCS16,
  author       = {Christopher B. Choy and
                  Danfei Xu and
                  JunYoung Gwak and
                  Kevin Chen and
                  Silvio Savarese},
  title        = {{3D-R2N2}: A Unified Approach for Single and Multi-view {3D} Object
                  Reconstruction},
  booktitle    = {European Conference Computer Vision},
  pages        = {628--644},
  year         = {2016}
}

@article{DBLP:journals/pami/TulsianiKCM17,
  author       = {Shubham Tulsiani and
                  Abhishek Kar and
                  Jo{\~{a}}o Carreira and
                  Jitendra Malik},
  title        = {Learning Category-Specific Deformable {3D} Models for Object Reconstruction},
  journal      = {{IEEE} Transactions on Pattern Analysis and Machine Intelligence},
  volume       = {39},
  number       = {4},
  pages        = {719--731},
  year         = {2017}
}

@inproceedings{DBLP:conf/cvpr/OswaldTC12,
  author       = {Martin R. Oswald and
                  Eno T{\"{o}}ppe and
                  Daniel Cremers},
  title        = {Fast and globally optimal single view reconstruction of curved objects},
  booktitle    = {{IEEE} Conference on Computer Vision and Pattern Recognition},
  pages        = {534--541},
  year         = {2012}
}

@inproceedings{DBLP:conf/nips/SaxenaCN05,
  author       = {Ashutosh Saxena and
                  Sung H. Chung and
                  Andrew Y. Ng},
  title        = {Learning Depth from Single Monocular Images},
  booktitle    = {Advances in Neural Information Processing Systems},
  year         = {2005}
}

@inproceedings{DBLP:conf/nips/LiLLZLLQZXLTWLG24,
  author       = {Peng Li and
                  Yuan Liu and
                  Xiaoxiao Long and
                  Feihu Zhang and
                  Cheng Lin and
                  Mengfei Li and
                  Xingqun Qi and
                  Shanghang Zhang and
                  Wei Xue and
                  Wenhan Luo and
                  Ping Tan and
                  Wenping Wang and
                  Qifeng Liu and
                  Yike Guo},
  title        = {{Era3D}: High-Resolution Multiview Diffusion using Efficient Row-wise
                  Attention},
  booktitle    = {Advances in Neural Information Processing Systems},
  year         = {2024}
}

@inproceedings{DBLP:conf/nips/DarasDDD23,
  author       = {Giannis Daras and
                  Yuval Dagan and
                  Alex Dimakis and
                  Constantinos Daskalakis},
  title        = {Consistent Diffusion Models: Mitigating Sampling Drift by Learning
                  to be Consistent},
  booktitle    = {Advances in Neural Information Processing Systems},
  year         = {2023}
}

@inproceedings{DBLP:conf/cvpr/YangHWGZZ0L24,
  author       = {Yunhan Yang and
                  Yukun Huang and
                  Xiaoyang Wu and
                  Yuan{-}Chen Guo and
                  Song{-}Hai Zhang and
                  Hengshuang Zhao and
                  Tong He and
                  Xihui Liu},
  title        = {{DreamComposer}: Controllable 3{D} Object Generation via Multi-View Conditions},
  booktitle    = {{IEEE/CVF} Conference on Computer Vision and Pattern Recognition},
  pages        = {8111--8120},
  year         = {2024}
}

@inproceedings{DBLP:conf/cvpr/CaoYL0X025,
  author       = {Chenjie Cao and
                  Chaohui Yu and
                  Shang Liu and
                  Fan Wang and
                  Xiangyang Xue and
                  Yanwei Fu},
  title        = {{MVGenMaster}: Scaling Multi-View Generation from Any Image via {3D} Priors
                  Enhanced Diffusion Model},
  booktitle    = {{IEEE/CVF} Conference on Computer Vision and Pattern Recognition},
  pages        = {6045--6056},
  year         = {2025}
}

@inproceedings{DBLP:conf/nips/WangLLTKW21,
  author       = {Peng Wang and
                  Lingjie Liu and
                  Yuan Liu and
                  Christian Theobalt and
                  Taku Komura and
                  Wenping Wang},
  title        = {{NeuS}: Learning Neural Implicit Surfaces by Volume Rendering for Multi-view
                  Reconstruction},
  booktitle    = {Advances in Neural Information Processing Systems},
  pages        = {27171--27183},
  year         = {2021}
}

@article{DBLP:journals/tvcg/HartwigESKPPGBR25,
  author       = {Sebastian Hartwig and
                  Dominik Engel and
                  Leon Sick and
                  Hannah Kniesel and
                  Tristan Payer and
                  Poonam Poonam and
                  Michael Gl{\"{o}}ckler and
                  Alex B{\"{a}}uerle and
                  Timo Ropinski},
  title        = {A Survey on Quality Metrics for Text-to-Image Generation},
  journal      = {{IEEE} Transactions on Visualization and Computer Graphics},
  volume       = {31},
  number       = {10},
  pages        = {9464--9483},
  year         = {2025}
}

@inproceedings{DBLP:conf/icml/ChenK0H20,
  author       = {Ting Chen and
                  Simon Kornblith and
                  Mohammad Norouzi and
                  Geoffrey E. Hinton},
  title        = {A Simple Framework for Contrastive Learning of Visual Representations},
  booktitle    = {International Conference on Machine Learning},
  pages        = {1597--1607},
  year         = {2020}
}

@inproceedings{DBLP:conf/cvpr/ShumHNY25,
  author       = {Ka{-}Chun Shum and
                  Binh{-}Son Hua and
                  Duc Thanh Nguyen and
                  Sai{-}Kit Yeung},
  title        = {Color Alignment in Diffusion},
  booktitle    = {{IEEE/CVF} Conference on Computer Vision and Pattern Recognition},
  pages        = {28446--28455},
  year         = {2025}
}

@inproceedings{DBLP:conf/cvpr/RombachBLEO22,
  author       = {Robin Rombach and
                  Andreas Blattmann and
                  Dominik Lorenz and
                  Patrick Esser and
                  Bj{\"{o}}rn Ommer},
  title        = {High-Resolution Image Synthesis with Latent Diffusion Models},
  booktitle    = {{IEEE/CVF} Conference on Computer Vision and Pattern Recognition},
  pages        = {10674--10685},
  year         = {2022}
}

@inproceedings{ruiz_dreambooth_2023,
	title = {{DreamBooth}: Fine Tuning Text-to-Image Diffusion Models for Subject-Driven Generation},
	booktitle = {{IEEE}/{CVF} Conference on Computer Vision and Pattern Recognition},
	pages = {22500--22510},
	author = {Ruiz, Nataniel and Li, Yuanzhen and Jampani, Varun and Pritch, Yael and Rubinstein, Michael and Aberman, Kfir},
	year = {2023}
}

@inproceedings{ho_denoising_2020,
	title = {Denoising Diffusion Probabilistic Models},
	pages = {6840--6851},
	booktitle = {Advances in Neural Information Processing Systems},
	author = {Ho, Jonathan and Jain, Ajay and Abbeel, Pieter},
	year = {2020},
}

@inproceedings{lin_magic3d_2023_1,
  author       = {Chen{-}Hsuan Lin and
                  Jun Gao and
                  Luming Tang and
                  Towaki Takikawa and
                  Xiaohui Zeng and
                  Xun Huang and
                  Karsten Kreis and
                  Sanja Fidler and
                  Ming{-}Yu Liu and
                  Tsung{-}Yi Lin},
  title        = {Magic3{D}: High-Resolution {T}ext-to-3{D} Content Creation},
  booktitle    = {{IEEE/CVF} Conference on Computer Vision and Pattern Recognition},
  pages        = {300--309},
  year         = {2023}
}

@inproceedings{raj_dreambooth3D_2023,
	title = {{DreamBooth3D}: Subject-Driven Text-to-3{D} Generation},
	pages = {2349--2359},
	booktitle = {{IEEE/CVF} International Conference on Computer Vision},
    year={2023},
	author = {Raj, Amit and Kaza, Srinivas and Poole, Ben and Niemeyer, Michael and Ruiz, Nataniel and Mildenhall, Ben and Zada, Shiran and Aberman, Kfir and Rubinstein, Michael and Barron, Jonathan and Li, Yuanzhen and Jampani, Varun}
}

@inproceedings{long_wonder3D_2024,
  author       = {Xiaoxiao Long and
                  Yuan{-}Chen Guo and
                  Cheng Lin and
                  Yuan Liu and
                  Zhiyang Dou and
                  Lingjie Liu and
                  Yuexin Ma and
                  Song{-}Hai Zhang and
                  Marc Habermann and
                  Christian Theobalt and
                  Wenping Wang},
  title        = {Wonder3{D}: Single Image to 3{D} Using Cross-Domain Diffusion},
  booktitle    = {{IEEE/CVF} Conference on Computer Vision and Pattern Recognition},
  pages        = {9970--9980},
  year         = {2024}
}

@inproceedings{saharia_photorealistic_2022,
  author       = {Chitwan Saharia and
                  William Chan and
                  Saurabh Saxena and
                  Lala Li and
                  Jay Whang and
                  Emily L. Denton and
                  Seyed Kamyar Seyed Ghasemipour and
                  Raphael Gontijo Lopes and
                  Burcu Karagol Ayan and
                  Tim Salimans and
                  Jonathan Ho and
                  David J. Fleet and
                  Mohammad Norouzi},
  title        = {Photorealistic Text-to-Image Diffusion Models with Deep Language Understanding},
  booktitle    = {Advances in Neural Information Processing Systems},
  year         = {2022}
}

@inproceedings{liu_one-2-3-45_2023,
  author       = {Minghua Liu and
                  Ruoxi Shi and
                  Linghao Chen and
                  Zhuoyang Zhang and
                  Chao Xu and
                  Xinyue Wei and
                  Hansheng Chen and
                  Chong Zeng and
                  Jiayuan Gu and
                  Hao Su},
  title        = {One-2-3-45++: Fast Single Image to 3{D} Objects with Consistent Multi-View
                  Generation and 3{D} Diffusion},
  booktitle    = {{IEEE/CVF} Conference on Computer Vision and Pattern Recognition},
  pages        = {10072--10083},
  year         = {2024}
}

@inproceedings{DBLP:journals/corr/abs-2303-11328,
  author       = {Ruoshi Liu and
                  Rundi Wu and
                  Basile Van Hoorick and
                  Pavel Tokmakov and
                  Sergey Zakharov and
                  Carl Vondrick},
  title        = {Zero-1-to-3: Zero-shot One Image to 3{D} Object},
  booktitle      = {{IEEE/CVF} International Conference on Computer Vision},
  pages       = {9264--9275},
  year         = {2023}
}

@inproceedings{liu_syncdreamer_2024,
  author       = {Yuan Liu and
                  Cheng Lin and
                  Zijiao Zeng and
                  Xiaoxiao Long and
                  Lingjie Liu and
                  Taku Komura and
                  Wenping Wang},
  title        = {Sync{D}reamer: Generating Multiview-consistent Images from a Single-view
                  Image},
  booktitle    = {International Conference on Learning Representations},
  year         = {2024}
}

@inproceedings{RealFusion,
  author       = {Luke Melas{-}Kyriazi and
                  Christian Rupprecht and
                  Iro Laina and
                  Andrea Vedaldi},
  title        = {Real{F}usion: 360{\textdegree} Reconstruction of Any Object from a Single
                  Image},
  booktitle      = {{IEEE/CVF} Conference on Computer Vision and Pattern Recognition},
  year         = {2023},
  pages = {8446--8455}
}

@article{nichol2022pointe,
  author       = {Alex Nichol and
                  Heewoo Jun and
                  Prafulla Dhariwal and
                  Pamela Mishkin and
                  Mark Chen},
  title        = {Point-{E}: {A} System for Generating 3{D} Point Clouds from Complex Prompts},
  journal      = {CoRR},
  volume       = {abs/2212.08751},
  year         = {2022}
}

@article{jun2023shape,
  author       = {Heewoo Jun and
                  Alex Nichol},
  title        = {{Shap-E}: Generating Conditional 3{D} Implicit Functions},
journal      = {CoRR},
  volume       = {abs/2305.02463},
  year         = {2023},
}

@inproceedings{hong2023lrm,
  title        = {{LRM}: Large Reconstruction Model for Single Image to 3{D}},
  author={Hong, Yicong and Zhang, Kai and Gu, Jiuxiang and Bi, Sai and Zhou, Yang and Liu, Difan and Liu, Feng and Sunkavalli, Kalyan and Bui, Trung and Tan, Hao},
  year={2024},
  booktitle    = {International Conference on Learning Representations}
}

@article{xu2024instantmesh,
  author       = {Jiale Xu and
                  Weihao Cheng and
                  Yiming Gao and
                  Xintao Wang and
                  Shenghua Gao and
                  Ying Shan},
  title        = {Instant{M}esh: Efficient 3{D} Mesh Generation from a Single Image with
                  Sparse-view Large Reconstruction Models},
  journal      = {CoRR},
  volume       = {abs/2404.07191},
  year         = {2024},}

@inproceedings{wang2024crm,
  author       = {Zhengyi Wang and
                  Yikai Wang and
                  Yifei Chen and
                  Chendong Xiang and
                  Shuo Chen and
                  Dajiang Yu and
                  Chongxuan Li and
                  Hang Su and},
  title        = {{CRM:} Single Image to 3{D} Textured Mesh with Convolutional Reconstruction
                  Model},
  booktitle    = {European Conference on Computer Vision},
  pages        = {57--74},
  year         = {2024},
}

@inproceedings{downs_google_2022,
	author = {Downs, Laura and Francis, Anthony and Koenig, Nate and Kinman, Brandon and Hickman, Ryan and Reymann, Krista and {McHugh}, Thomas B. and Vanhoucke, Vincent},
    booktitle = {International Conference on Robotics and Automation},
    year = {2022},
    pages = {2553--2560},
    title = {Google {S}canned {O}bjects: {A} High-Quality Dataset of 3{D} Scanned Household Items}
}

@inproceedings{chen2023fantasia3D,
  author       = {Rui Chen and
                  Yongwei Chen and
                  Ningxin Jiao and
                  Kui Jia},
  title        = {Fantasia3{D}: Disentangling Geometry and Appearance for High-quality
                  Text-to-3{D} Content Creation},
  booktitle    = {{IEEE/CVF} International Conference on Computer Vision},
  pages        = {22189--22199},
  year         = {2023}
}

@article{Qin_2020_PR,
title = {U\({}^{\mbox{2}}\)-{N}et: Going deeper with nested {U}-structure for salient
                  object detection},
author = {Qin, Xuebin and Zhang, Zichen and Huang, Chenyang and Dehghan, Masood and Zaiane, Osmar and Jagersand, Martin},
journal = {Pattern Recognition},
volume = {106},
pages = {107404},
year = {2020}
}

@article{galesic2009effects,
  title={Effects of questionnaire length on participation and indicators of response quality in a web survey},
  author={Galesic, M. and Bosnjak, M.},
  journal={Public opinion quarterly},
  volume={73},
  number={2},
  pages={349--360},
  year={2009},
}

@InProceedings{Xu_2025_ICCV,
    author    = {Jiale Xu and
                  Shenghua Gao and
                  Ying Shan},
    title     = {{FreeSplatter}: Pose-free Gaussian Splatting for Sparse-view 3{D} Reconstruction},
    booktitle = {{IEEE/CVF} International Conference on Computer Vision},
    pages        = {25442--25452},
    year      = {2025}
}

@inproceedings{zhao2024sparseags,
  title={Sparse-view Pose Estimation and Reconstruction via Analysis by Generative Synthesis}, 
  author={Qitao Zhao and Shubham Tulsiani},
  booktitle={Advances in Neural Information Processing Systems},
  year={2024}
}

@article{DBLP:journals/corr/abs-2308-06721,
  author       = {Hu Ye and
                  Jun Zhang and
                  Sibo Liu and
                  Xiao Han and
                  Wei Yang},
  title        = {{IP}-{A}dapter: Text Compatible Image Prompt Adapter for Text-to-Image
                  Diffusion Models},
  journal      = {CoRR},
  year         = {2023},
  }

@inproceedings{DBLP:conf/iccv/ZhangRA23,
  author       = {Lvmin Zhang and
                  Anyi Rao and
                  Maneesh Agrawala},
  title        = {Adding Conditional Control to Text-to-Image Diffusion Models},
  booktitle    = {{IEEE/CVF} International Conference on Computer Vision},
  pages        = {3813--3824},
  year         = {2023},
}

\end{document}